%% file: atlas.tex
\documentclass[11pt,letterpaper]{article}
\usepackage[T1]{fontenc}
\usepackage[letterpaper,textwidth=6in,textheight=8.5in,centering]{geometry}
\usepackage{lmodern}
\usepackage{amsmath,amssymb,amsthm}
\usepackage{natbib}
\usepackage{booktabs,float,tikz,graphicx}
\usepackage[hidelinks]{hyperref}
\bibpunct{(}{)}{;}{a}{,}{,}
\newtheorem{proposition}{Proposition}
\newenvironment{keywords}{\par\small\noindent\textbf{Keywords:} }{\par}

\input{numbers_atlas.tex}

\input{numbers_identity.tex}

\input{numbers_extras.tex}

\input{numbers_revision.tex}
\input{numbers_library.tex}
\hypersetup{pdftitle={Auditing Identity and Timing Signals in Operational Logs},pdfauthor={Andrew Hendel}}
\title{Auditing Identity and Timing Signals in Operational Logs}
\author{Andrew Hendel\\Latent Archon LLC\\New York, NY, USA\\\texttt{ajhendel@latentarchon.com}}
\date{Preprint}
\begin{document}
\maketitle

\begin{abstract}%
Models that predict events from operational logs, such as sepsis in
ICU records or failures in hard drives, can score well by recognizing
which patient or drive a record came from, or when it was recorded,
instead of predicting the event. I introduce two simple audits that
measure how much of this information a dataset carries before any
event-prediction model is trained, and apply them to seven public
datasets in six domains. One asks whether a classifier can tell that two records came
from the same unit; the other asks which of two records from the same
unit came later. Under the stated nulls, both have expected AUC 0.5. In ICU
data, the identity audit flags a common preprocessing step: carrying
the last measurement forward raises identity AUC from \RvIdRawAuc{} to
\RvIdLocfAuc{}. Discrimination persists against feature-similar
controls: when the different unit is drawn from those most similar to
the first, identity AUC stays at \NrMarFullAuc{} or above for drives,
mortgages, vessels and carried-forward ICU records. On a 12-hour sepsis forecasting task, carrying forward makes
a model look \RvFcLocfRawOvLead{} AUC better than raw values when a
patient's hours are split between training and test, but only
\RvFcLocfRawDjLead{} better when patients are held out. The
exaggeration under patient overlap also appears in a second ICU
dataset with mortality as the outcome. On the sepsis task, a
chronological control with complete follow-up, in which training sees
only a patient's earlier, resolved hours, shows no positive mean
overlap gap. Timing is a strong
predictor: with complete follow-up, hour of stay and measurement counts
alone, with no clinical value, score \RvChCountsTimeUnitDj{} on
held-out patients, and adding the raw clinical values changes that by
\RvChLeadRawTimeCountsTimeDj{}. Outside the ICU, \BbStraddle{} of
drives and \FmStraddle{} of loans have records on both sides of a
calendar cut, and the overlap gap persists under a calendar split. The
audits show which features and
preprocessing choices to check; the evaluation protocol decides the
consequences. I release the code, a library that runs the audits on
any table with a unit column and a time column, and a guide to using
the audits.
\end{abstract}

\begin{keywords}
  evaluation, data leakage, imputation, dataset audits, negative
  controls, operational logs, clinical time series
\end{keywords}

\section{Introduction}
\label{sec:intro}

An event-prediction model may recognize the patient, loan or drive
that produced a record, or infer where that record falls in time.
When labels are correlated within units, recognizing a unit seen in
training can improve test scores without improving predictions for
new units. Prior work documents these evaluation failures
\citep{kaufman2012leakage,kapoor2023leakage,tampu2022inflation,neto2019identity}.
The appropriate split follows from deployment: new units require
unit-disjoint evaluation, while predictions for known units require
training information available before prediction. An audit score
does not change those requirements.

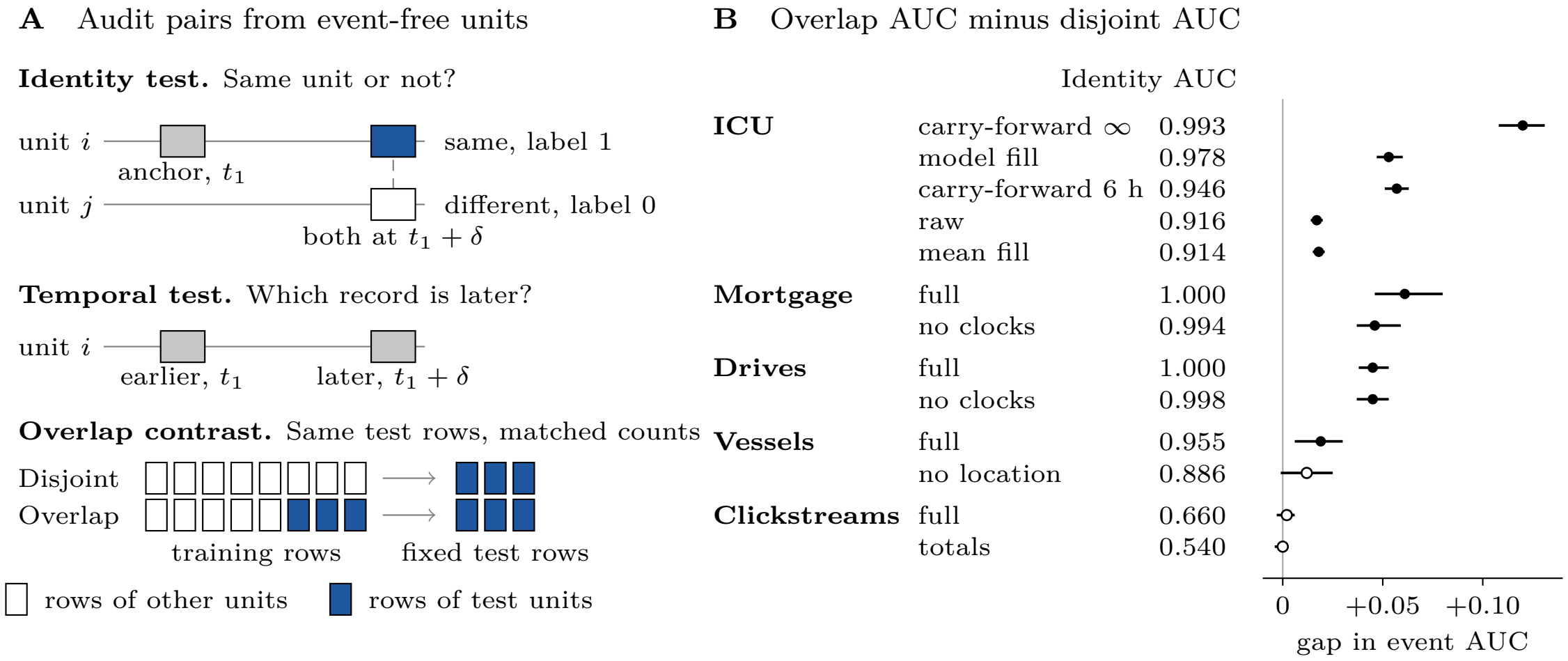
\begin{figure}[t]
\centering
\definecolor{unitblue}{RGB}{31,88,160}
\newcommand{\fxo}{5.55}
\newcommand{\fxs}{19}
\newcommand{\frow}[6]{
  \node[anchor=west] at (1.95,#1) {#2};
  \node[anchor=east] at (5.15,#1) {#3};
  \pgfmathsetmacro\xd{\fxo+(0#4)*\fxs}
  \pgfmathsetmacro\xl{\fxo+(0#5)*\fxs}
  \pgfmathsetmacro\xh{\fxo+(0#6)*\fxs}
  \pgfmathtruncatemacro\excl{(0#5)>0}
  \draw[line width=0.7pt] (\xl,#1)--(\xh,#1);
  \ifnum\excl=1
    \fill (\xd,#1) circle (1.4pt);
  \else
    \draw[fill=white,line width=0.5pt] (\xd,#1) circle (1.4pt);
  \fi}
\begin{tikzpicture}[x=1cm,y=1cm,font=\fontsize{7}{8.4}\selectfont,
  every node/.style={text height=1.4ex,text depth=0.4ex},
  rec/.style={draw,minimum width=0.42cm,minimum height=0.24cm,inner sep=0pt},
  sq/.style={draw,minimum size=0.2cm,inner sep=0pt}]
\node[anchor=west,font=\fontsize{8}{9.6}\selectfont] at (0,5.6) {\textbf{A}\quad Audit pairs from event-free units};
\node[anchor=west] at (0,5.05) {\textbf{Identity test.} Same unit or not?};
\node[anchor=west] at (0,4.45) {unit $i$};
\node[anchor=west] at (0,3.85) {unit $j$};
\draw[gray] (0.95,4.45)--(4.0,4.45);
\draw[gray] (0.95,3.85)--(4.0,3.85);
\draw[densely dashed,gray] (3.7,4.45)--(3.7,3.85);
\node[rec,fill=gray!45] at (1.7,4.45) {};
\node[rec,fill=unitblue] at (3.7,4.45) {};
\node[rec,fill=white] at (3.7,3.85) {};
\node at (1.7,4.17) {anchor, $t_1$};
\node[anchor=west] at (4.05,4.45) {same, label 1};
\node[anchor=west] at (4.05,3.85) {different, label 0};
\node at (3.7,3.55) {both at $t_1+\delta$};
\node[anchor=west] at (0,3.0) {\textbf{Temporal test.} Which record is later?};
\node[anchor=west] at (0,2.5) {unit $i$};
\draw[gray] (0.95,2.5)--(4.0,2.5);
\node[rec,fill=gray!45] at (1.7,2.5) {};
\node[rec,fill=gray!45] at (3.7,2.5) {};
\node at (1.7,2.22) {earlier, $t_1$};
\node at (3.7,2.22) {later, $t_1+\delta$};
\node[anchor=west] at (0,1.7) {\textbf{Overlap contrast.} Same test rows, matched counts};
\node[anchor=west] at (0,1.25) {Disjoint};
\node[anchor=west] at (0,0.9) {Overlap};
\foreach \k in {0,...,7} {\node[sq,fill=white] at (1.45+0.27*\k,1.25) {};}
\foreach \k in {0,...,4} {\node[sq,fill=white] at (1.45+0.27*\k,0.9) {};}
\foreach \k in {5,...,7} {\node[sq,fill=unitblue] at (1.45+0.27*\k,0.9) {};}
\foreach \y in {1.25,0.9} {
  \draw[->,gray] (3.6,\y)--(4.1,\y);
  \foreach \k in {0,...,2} {\node[sq,fill=unitblue] at (4.4+0.27*\k,\y) {};}}
\node at (2.4,0.55) {training rows};
\node at (4.67,0.55) {fixed test rows};
\node[sq,fill=white] at (0.12,0.1) {};
\node[anchor=west] at (0.25,0.1) {rows of other units};
\node[sq,fill=unitblue] at (3.2,0.1) {};
\node[anchor=west] at (3.33,0.1) {rows of test units};
\begin{scope}[xshift=6.6cm]
\node[anchor=west,font=\fontsize{8}{9.6}\selectfont] at (0,5.6) {\textbf{B}\quad Overlap AUC minus disjoint AUC};
\node[anchor=east] at (5.25,5.05) {Identity AUC};
\draw[gray!70] (\fxo,0.3)--(\fxo,4.85);
\draw (\fxo-0.01*\fxs,0.3)--(\fxo+0.14*\fxs,0.3);
\foreach \v/\l in {0/0,0.05/{+0.05},0.10/{+0.10}} {
  \draw (\fxo+\v*\fxs,0.3)--(\fxo+\v*\fxs,0.23);
  \node at (\fxo+\v*\fxs,0.05) {\l};}
\node at (\fxo+0.065*\fxs,-0.3) {gap in event AUC};
\node[anchor=west] at (0,4.6) {\textbf{ICU}};
\frow{4.6}{carry-forward $\infty$}{\RvIdLocfAuc}{\RvIcuLocfUnitDelta}{\RvIcuLocfUnitLo}{\RvIcuLocfUnitHi}
\frow{4.3}{model fill}{\RvIdModelAuc}{\RvIcuModelTrainUnitDelta}{\RvIcuModelTrainUnitLo}{\RvIcuModelTrainUnitHi}
\frow{4.0}{carry-forward 6 h}{\RvIdLocfSixAuc}{\RvIcuLocfSixUnitDelta}{\RvIcuLocfSixUnitLo}{\RvIcuLocfSixUnitHi}
\frow{3.7}{raw}{\RvIdRawAuc}{\RvIcuRawUnitDelta}{\RvIcuRawUnitLo}{\RvIcuRawUnitHi}
\frow{3.4}{mean fill}{\RvIdMeanAuc}{\RvIcuMeanTrainUnitDelta}{\RvIcuMeanTrainUnitLo}{\RvIcuMeanTrainUnitHi}
\node[anchor=west] at (0,3.0) {\textbf{Mortgage}};
\frow{3.0}{full}{\IdFmFullAuc}{\RvFmFullUnitDelta}{\RvFmFullUnitLo}{\RvFmFullUnitHi}
\frow{2.7}{no clocks}{\IdFmNoClocksAuc}{\RvFmNoClocksUnitDelta}{\RvFmNoClocksUnitLo}{\RvFmNoClocksUnitHi}
\node[anchor=west] at (0,2.3) {\textbf{Drives}};
\frow{2.3}{full}{\IdBbFullAuc}{\RvBbFullUnitDelta}{\RvBbFullUnitLo}{\RvBbFullUnitHi}
\frow{2.0}{no clocks}{\IdBbNoClocksAuc}{\RvBbNoClocksUnitDelta}{\RvBbNoClocksUnitLo}{\RvBbNoClocksUnitHi}
\node[anchor=west] at (0,1.6) {\textbf{Vessels}};
\frow{1.6}{full}{\IdMarFullAuc}{\RvMarFullUnitDelta}{\RvMarFullUnitLo}{\RvMarFullUnitHi}
\frow{1.3}{no location}{\IdMarNoLocAuc}{\RvMarNoLocUnitDelta}{\RvMarNoLocUnitLo}{\RvMarNoLocUnitHi}
\node[anchor=west] at (0,0.9) {\textbf{Clickstreams}};
\frow{0.9}{full}{\IdOuladFullAuc}{\RvOuladFullUnitDelta}{\RvOuladFullUnitLo}{\RvOuladFullUnitHi}
\frow{0.6}{totals}{\IdOuladTotalsAuc}{\RvOuladTotalsUnitDelta}{\RvOuladTotalsUnitLo}{\RvOuladTotalsUnitHi}
\end{scope}
\end{tikzpicture}
\caption{The two audits and what they flag. (A) Both audits pair
records of event-free units. The overlap contrast scores the same test
rows after training without or with other rows of the test units, at
matched per-class training counts. (B) Overlap AUC minus disjoint AUC
for the \RvSepCells{} rows of Table~\ref{tab:bridge}, with each row's
identity AUC. Points are means over 200 replicates, bars are 2.5--97.5
percentile ranges, and a filled point has a range that excludes zero.
In the ICU, unlimited carry-forward ($\infty$) raises the gap from
\RvIcuRawUnitDelta{} on raw values to \RvIcuLocfUnitDelta{}.
Clickstream gaps are near zero, with ranges that include zero.}
\label{fig:overview}
\end{figure}

I introduce two audits, both run on units with no recorded event
(Figure~\ref{fig:overview}A). The identity
test distinguishes same-unit from different-unit pairs matched on
time and offset. The temporal test distinguishes earlier from later
records within a unit. Using established classification and
verification methods
\citep{lopezpaz2017revisiting,jain2004introduction}, I measure unit
identity and temporal ordering as dataset properties, without fitting
a downstream event-prediction model and with expected held-out AUC 0.5
under the stated nulls. To my knowledge, no earlier work measures
both signals together on the same panels across domains. Controlled
downstream comparisons then examine how their consequences depend on
preprocessing and evaluation design.

What the audit adds is a measurement of which features and
preprocessing choices make records recognizable, and that measurement
points at real problems. In ICU data, carry-forward imputation makes
records from the same patient much easier to match (identity AUC
\RvIdRawAuc{} to \RvIdLocfAuc{}), and it exaggerates its own benefit
more than threefold when a patient's hours are randomly split across
training and test (\RvFcLocfRawOvLead{} against
\RvFcLocfRawDjLead{}). The overlap effect also appears in a second
ICU dataset with mortality as the outcome. On the sepsis task, a
within-patient chronological control with complete follow-up shows no
positive mean overlap gap, so a recognizable patient does not produce
an inflated score under every protocol. Timing works the same way: the audit locates timing signal in
ICU measurement counts, and on the sepsis task evaluated here hour of stay plus those
counts, with no clinical value, scores \RvChCountsTimeUnitDj{} on
held-out patients. Adding the raw clinical values raises that to
\RvChRawTimeUnitDj{} (complete follow-up). A calendar split does not
remove the unit: \BbStraddle{} of drives and \FmStraddle{} of loans
have records on both sides of an 80th-percentile cut, and the overlap
gap persists there (\RvBbFullTimeDelta{} for drives,
\RvFmFullTimeDelta{} for mortgages).

\textbf{Contributions.}
\begin{enumerate}
\item \emph{Controlled evaluation experiments.} Fixed test rows and
matched training counts isolate the effect of replacing other-unit
training rows with test-unit rows. The experiments cover six datasets
in five of the six domains, including both ICU datasets, and add
multiple imputations and sepsis forecasting with chronological and
clock-baseline controls
(Sections~\ref{sec:bridge}--\ref{sec:forecast}).
\item \emph{A survey of seven datasets in six domains.} Feature ablations locate
identity and timing information in operational logs. Mortgage clocks
account for most timing discrimination; ICU measurement counts
carry both signals. Drawing the different unit from those most
similar to the first shows that the identity signal belongs to
individual units in drives, mortgages, vessels and ICU records, and
mostly to the kind of student in clickstreams
(Section~\ref{sec:survey}).
\item \emph{Reusable software and practical guidance.} A panel
adapter and a declaration of units, timestamps, features and offsets
are enough to run the audits. Section~\ref{sec:using} explains how to
use them to choose follow-up comparisons, and
Appendix~\ref{app:harness} provides a demonstration requiring no
data download. A separate library, \texttt{whowhen}, runs the audits
on any table from one call (Appendix~\ref{app:library}).
\end{enumerate}

\section{The Two Tests}
\label{sec:definition}

Identified units $i$ produce feature rows $x_{i,t}$. Fix an offset
range $[\delta_{\min}, \delta_{\max}]$ suited to the domain's sampling
interval; both tests use the same range per dataset.

\textbf{Identity test.} Take a record, pair it
with either a later record from the same unit or a record from a
different unit at the same moment, and ask a classifier which kind of
pair it is. If nothing in the records identifies the unit, it can only
guess. Formally, sample an anchor $(i, t_1)$ and
offset $\delta$; emit $(x_{i,t_1}, x_{i,t_1+\delta})$ with label 1
and, for the same anchor and the same $t_1{+}\delta$,
$(x_{i,t_1}, x_{j,t_1+\delta})$ for a random $j \neq i$ observed at
that time, with label 0. Both classes share the anchor distribution,
the calendar, and the offset; only identity differs. Features are
$(x_a, x_b, |x_a - x_b|)$. Units are split before any pair is drawn,
and anchor and impostor come from the same side, so the split breaks
no matched pair. Intervals come from a unit-level bootstrap
(Appendix~\ref{app:validation}). The null is that no
persistent unit-specific dependence remains once calendar and offset
are matched, $(X_{i,t_1}, X_{i,t_1+\delta}) \stackrel{d}{=}
(X_{i,t_1}, X_{j,t_1+\delta})$ under the matching controls. Under it
the two classes are identically distributed.

\textbf{Temporal test.} Take two records from
the same unit, hours or days apart, and ask a classifier which one came
first. If nothing in the records changes with time, it can only guess.
Formally, a pair is $(x_{i,t_1}, x_{i,t_2})$,
$t_2 - t_1 \in [\delta_{\min}, \delta_{\max}]$, drawn from a unit
with no recorded adverse event; rows are labeled by temporal position
alone. A classifier trains on 80\% of units and is evaluated on
held-out units. The temporal AUC is the pooled AUC with a
unit-clustered bootstrap interval. The null is exchangeability of a
unit's ordinary windows across the offset range,
$(X_{i,t}, X_{i,t+\delta}) \stackrel{d}{=}
(X_{i,t+\delta}, X_{i,t})$ for
$\delta \in [\delta_{\min}, \delta_{\max}]$ under the declared
pairing and weighting. Under it the labels are a symmetric
convention and the expected AUC is 0.5 for any scorer.

\textbf{Under the null, expected empirical AUC is exactly 0.5.} If
exchanging the two items of every matched test pair leaves the joint
law of the test pairs unchanged, the pooled empirical AUC with
mid-rank ties has expectation one half, for any scorer fit without the
held-out labels and any number of pairs
(Proposition~\ref{prop:forced-null}, Appendix~\ref{app:proof}). The
swap invariance is the null hypothesis itself. For dependent matched
pairs it is stronger than equal class distributions, and the
unit-disjoint split does not establish it: the split only keeps the
scorer from being fit on the held-out units, which the result also
requires.

\textbf{Each test controls for the other property.} In the identity
test, matching on time and offset controls those coordinates when
comparing same-unit and different-unit pairs. The match is on the panel's time
coordinate, which for the ICU is hours since admission; matching on
hospital as well is a separate control (Appendix~\ref{app:extra-arms}). The identity
test uses feature values only; identity carried by \emph{when} a unit
is observed is outside it. In the temporal test both windows come from
one unit, so unit identity does not separate the labels. A cross-unit
placebo (Appendix~\ref{sec:placebo}) measures how much of the temporal
AUC is structure shared across units.

\textbf{What the numbers are.} Both audits use only units with no
recorded event, and neither classifier is fit to an event outcome.
Each AUC comes from one fixed configuration of LightGBM gradient
boosting \citep{ke2017lightgbm}, varied in Appendix~\ref{app:scorer}, so it is a lower bound on how
well the pairing task can be solved, and it depends on the declared
offsets and features. It is not an event-prediction score and cannot be
subtracted from, or compared with, a detector's AUC. The value 0.5 is a
reference point, and a score of 0.5 does not establish that an
evaluation is valid. Rows are scored pooled across the test set. I
report each quantity with all features, with declared clock columns
removed, and for a single channel, as means over five pair-draw seeds
with unit-level bootstrap intervals (Appendix~\ref{app:validation}).

\subsection{Eligibility}
\label{sec:eligibility}

Both tests require five things: (1)~identified units observed
repeatedly; (2)~enough units for a unit-disjoint split (tens of
thousands is comfortable; twenty works with wide intervals for the
temporal test and fails for the identity test,
Appendix~\ref{app:domains}); (3)~per-unit spans
longer than the offset range; (4)~features at each timestamp;
(5)~a clock declaration from the analyst. They do not require a
prediction task, labeled events beyond the column that excludes
event-bearing units, completeness, or any particular sampling rate.
For example, two public maintenance datasets do not qualify: MetroPT
has one unit, and Scania APS has one snapshot per truck.

\section{Identity and Timing Signals Across Six Domains}
\label{sec:survey}
\subsection{Identity Signal}
\label{sec:id-domains}

Table~\ref{tab:exposure} reports the identity survey. Each domain
follows the same protocol: an adapter maps raw data to the panel
format, a declaration fixes offsets, feature sets (tiers) and clocks, and the
measurement runs. Appendix~\ref{app:domains} describes each domain
and gives the drive, vessel and wind results in full. Drive identity
survives restricting impostors to the anchor's own drive model, and
vessel identity survives removing position. Clickstreams, near chance
on time (Section~\ref{sec:domains}), still give identity AUC
\IdOuladFullAuc{}.

\begin{table}[H]
\centering
\small
\begin{tabular}{llcc}
\toprule
Domain & Offsets & All features & Fewer features \\
\midrule
Drives & 30--90 d & \IdBbFullAuc{} & \IdBbNoClocksAuc{} (no clocks) \\
Mortgage & 6--18 mo & \IdFmFullAuc{} & \IdFmNoClocksAuc{} (no clocks) \\
ICU (carry-forward) & 6--24 h & \IdIcuFullAuc{} & \IdIcuTextureAuc{} (counts only) \\
Vessels & 14--60 d & \IdMarFullAuc{} & \IdMarNoLocAuc{} (no location) \\
ICU (no carry-forward) & 6--24 h & \RvIdRawAuc{} & \IdIcuRawFullAuc{} (values only) \\
ICU, PhysioNet 2012 (carry-forward) & 6--24 h & \RvIdIcuBLocfAuc{} & \RvIdIcuBCountsAuc{} (counts only) \\
ICU, PhysioNet 2012 (no carry-forward) & 6--24 h & \RvIdIcuBRawAuc{} & n/a \\
Clickstreams & 14--60 d & \IdOuladFullAuc{} & \IdOuladTotalsAuc{} (totals only) \\
Wind & 14--60 d & n/a & n/a \\
\bottomrule
\end{tabular}
\caption{Identity AUCs (five-seed means). The last column repeats
the test on the named subset of the features: without the clock
columns, without location, on the ICU measurement counts or raw values
alone, or on a student's total clicks and sites visited without the
per-activity breakdown. The ICU rows of each dataset include the
six-hour measurement counts and differ only in carry-forward. n/a
marks a subset that was not run and, for wind, a test that twenty
units cannot support.}
\label{tab:exposure}
\end{table}

\textbf{Measurement counts alone identify patients.} How often each
ICU channel was measured in the last six hours, with no physiological
values, gives identity AUC \IdIcuTextureAuc{} [\IdIcuTextureCiLo,
\IdIcuTextureCiHi]. Raw values alone give \IdIcuRawFullAuc{}
[\IdIcuRawFullCiLo, \IdIcuRawFullCiHi], consistent with physiological
biometrics \citep{odinaka2012ecg}. Care patterns are known to identify
patients \citep{johnson2019reidentification}; my measurement differs in
unit and purpose (Appendix~\ref{sec:relatedfull}). The same counts also
carry temporal position (\IcuTextureAuc{}), which extends the finding
that observation patterns encode the care process
\citep{agniel2018biases}.

\textbf{Carry-forward imputation raises identity AUC.} Each imputation
variant keeps the measurement counts and differs only in
how missing values are filled. Raw values plus counts give
\RvIdRawAuc{}, and carrying the last value forward gives
\RvIdLocfAuc{}. Carried-forward values recur across a patient's windows, providing
additional information for linking records. The same choice leaves the
temporal AUC almost unchanged, so preprocessing belongs inside the
audit.

\textbf{Discrimination persists against feature-similar controls.} A
high identity AUC could mean only that the two units of a
different-unit pair are of different kinds: a tug and a tanker, a
frail patient and a robust one. To assess this explanation I repeat
the test with the different unit drawn from the units most similar to
the first. Each unit is summarized by its mean of every audited
feature, standardized across units, and the different unit is drawn
from the five nearest among up to 256 units observed at the matched
time. Similarity is computed within each side of the unit split.
Table~\ref{tab:nearest} gives the result. Drives, mortgages, vessels
and carried-forward ICU records stay at \NrMarFullAuc{} or above,
providing evidence of discrimination beyond this control for unit type.
This does not rule out all group-level explanations. ICU records without
carry-forward fall from \RvIdRawAuc{} to \NrIcuVarRawAuc{}, and the
measurement counts alone from \IdIcuTextureAuc{} to
\NrIcuTextureAuc{}: part of that signal is the kind of patient and
part is the individual. Clickstreams fall from
\IdOuladFullAuc{} to \NrOuladFullAuc{}, close to chance, so what
little identity signal clickstreams carry is mostly the kind of
student.

\begin{table}[H]
\centering
\small
\begin{tabular}{lcc}
\toprule
Domain & Any other unit & Most similar unit [95\% interval] \\
\midrule
Drives & \IdBbFullAuc{} & \NrBbFullAuc{} [\NrBbFullCiLo, \NrBbFullCiHi] \\
Mortgage & \IdFmFullAuc{} & \NrFmFullAuc{} [\NrFmFullCiLo, \NrFmFullCiHi] \\
ICU (carry-forward) & \IdIcuFullAuc{} & \NrIcuFullAuc{} [\NrIcuFullCiLo, \NrIcuFullCiHi] \\
Vessels & \IdMarFullAuc{} & \NrMarFullAuc{} [\NrMarFullCiLo, \NrMarFullCiHi] \\
ICU (no carry-forward) & \RvIdRawAuc{} & \NrIcuVarRawAuc{} [\NrIcuVarRawCiLo, \NrIcuVarRawCiHi] \\
Clickstreams & \IdOuladFullAuc{} & \NrOuladFullAuc{} [\NrOuladFullCiLo, \NrOuladFullCiHi] \\
\bottomrule
\end{tabular}
\caption{Identity AUC on all features when the different unit is any
other unit observed at the matched time (Table~\ref{tab:exposure}) and
when it is one of the five units most similar to the first. Five-seed
means; intervals pool the unit bootstrap over seeds. Offsets, pairs
and classifier are those of Table~\ref{tab:exposure}.}
\label{tab:nearest}
\end{table}

\subsection{Timing Signal}
\label{sec:domains}

Table~\ref{tab:floors} reports the temporal survey, on the same
panels and offset ranges as the identity survey.

\begin{table}[H]
\centering
\small
\begin{tabular}{llccc}
\toprule
Domain & Offsets & All features & Fewer features & Single channel \\
\midrule
Mortgage & 6--18 mo & \FmFullAuc{} & \FmNoClocksAuc{} & \FmRateAuc{} (rate) \\
ICU & 6--24 h & \IcuFullAuc{} & \IcuValuesAuc{} & \IcuTextureAuc{} (counts) \\
ICU (PhysioNet 2012) & 6--24 h & \IcuBFullAuc{} & \IcuBValuesAuc{} & \IcuBTextureAuc{} (counts) \\
Drives & 30--90 d & \BbFullAuc{} & \BbNoClocksAuc{} & \BbTempAuc{} (temp.) \\
Wind & 14--60 d & \WindFullAuc{} & \WindNoClocksAuc{} & \WindAmbientAuc{} (ambient) \\
Clickstreams & 14--60 d & \OuladFullAuc{} & \OuladTotalsAuc{} & n/a \\
Vessels & 14--60 d & \MarFullAuc{} & \MarNoLocAuc{} & \MarKinAuc{} (kinematics) \\
\bottomrule
\end{tabular}
\caption{Temporal AUCs in six operational-log domains (five-seed
means). ``Fewer features'' uses the subsets of
Table~\ref{tab:exposure} (ICU: raw values without counts).}
\label{tab:floors}
\end{table}

\textbf{Mortgages span nearly the whole range.} With amortization
clocks (balance, age, remaining term) the temporal AUC is
\FmFullAuc{}, because a vintage's loans amortize in lockstep. Without
them it drops to \FmNoClocksAuc{}. The loan's fixed rate alone gives
\FmRateAuc{}, the value symmetry forces for a feature that is constant
within a unit.

\textbf{ICU temporal AUC is high and does not come from clock columns.}
It is \IcuFullAuc{} [\IcuFullCiLo, \IcuFullCiHi] on never-septic
stays. Values alone give \IcuValuesAuc{}, and measurement counts
alone, with no physiological values, give \IcuTextureAuc{}: how often
each channel is measured changes over a stay. A cross-unit placebo
(Appendix~\ref{sec:placebo}) reproduces most of it, so most of this is
structure shared across patients at matched stay hours. The within-unit AUC contrast is
only \PlcIcuMidDiff{} at these offsets. Section~\ref{sec:forecast}
shows a consequence: the hour of stay alone outscores raw values that omit it
on sepsis forecasting. The second ICU dataset shows the same pattern,
with measurement counts alone at \IcuBTextureAuc{} [\IcuBTextureCiLo,
\IcuBTextureCiHi].

\textbf{Three domains are near chance.} Wind SCADA gives
\WindFullAuc{} (weather decorrelates within days), student
clickstreams give \OuladFullAuc{} (on active days the mix of activity
is nearly stationary; the calendar mainly decides which days are
active), and casualty-free vessel tracks give \MarFullAuc{}
[\MarFullCiLo, \MarFullCiHi], or \MarNoLocAuc{} without position. A
linear classifier gives the same within-domain conclusions with one
exception, the order of two ICU tiers (Appendix~\ref{app:scorer}), and Appendix~\ref{sec:offsets} shows how
the temporal AUC grows with the time offset.

\section{From Audit Scores to Inflated Evaluation}
\label{sec:bridge}

The audit measures information detectable by a specified probe.
This section tests sensitivity to overlap under controlled evaluation
designs. I start with unit-labeled case-control
classification, in which every row of an event unit is positive and
every row of a sampled non-event unit is negative. Section~\ref{sec:forecast}
tests the same comparison on a fixed-horizon forecasting task.
Wind has no event labels and is absent. For vessels the event is a
Coast Guard casualty.

\textbf{Design.} Each overlap contrast in Table~\ref{tab:bridge} scores
the same test rows with matched per-class training counts. About 20\% of units are selected
for testing; each of their sampled rows is independently assigned to
test with probability one half. Overlap training replaces some
remaining-unit rows with other rows of test units, matching class
counts. Disjoint training uses only remaining units.
The calendar contrast tests beyond the 80th time percentile and trains
at or before it; the embargo tests beyond the 90th and trains at or
before the 70th. $\Delta$ is overlap AUC minus disjoint AUC, a protocol
contrast rather than a decomposition of recognition alone.

Each cell averages 200 replicates, sampling up to 4{,}000 units per
class from fixed supersets and up to 20 rows per unit from fixed
reservoirs capped at 100 rows. A class with no surplus units retains
all its units. Intervals are the empirical 2.5--97.5 percentile ranges
over pool, row, split and fit variation, conditional on those
supersets; they are not confidence intervals for a population mean.
Tiers share each replicate.

A separate training-size control changes AUC by at most
\RvSizeMax{} (Appendix~\ref{app:bridge-submitted}).

\begin{table}[H]
\centering
\footnotesize
\setlength{\tabcolsep}{3pt}
\begin{tabular}{llcccccc}
\toprule
Domain & Tier & Identity AUC & Overlap & Disjoint & $\Delta$ unit & $\Delta$ calendar & $\Delta$ embargo \\
\midrule
ICU & carry-forward & \RvIdLocfAuc{} & \RvIcuLocfUnitOv{} & \RvIcuLocfUnitDj{} & \RvIcuLocfUnitDelta{} & \RvIcuLocfTimeDelta{} & \RvIcuLocfGapDelta{} \\
ICU & carry-forward 6 h & \RvIdLocfSixAuc{} & \RvIcuLocfSixUnitOv{} & \RvIcuLocfSixUnitDj{} & \RvIcuLocfSixUnitDelta{} & \RvIcuLocfSixTimeDelta{} & \RvIcuLocfSixGapDelta{} \\
ICU & model fill & \RvIdModelAuc{} & \RvIcuModelTrainUnitOv{} & \RvIcuModelTrainUnitDj{} & \RvIcuModelTrainUnitDelta{} & \RvIcuModelTrainTimeDelta{} & \RvIcuModelTrainGapDelta{} \\
ICU & raw & \RvIdRawAuc{} & \RvIcuRawUnitOv{} & \RvIcuRawUnitDj{} & \RvIcuRawUnitDelta{} & \RvIcuRawTimeDelta{} & \RvIcuRawGapDelta{} \\
ICU & mean fill & \RvIdMeanAuc{} & \RvIcuMeanTrainUnitOv{} & \RvIcuMeanTrainUnitDj{} & \RvIcuMeanTrainUnitDelta{} & \RvIcuMeanTrainTimeDelta{} & \RvIcuMeanTrainGapDelta{} \\
Mortgage & full & \IdFmFullAuc{} & \RvFmFullUnitOv{} & \RvFmFullUnitDj{} & \RvFmFullUnitDelta{} & \RvFmFullTimeDelta{} & \RvFmFullGapDelta{} \\
Mortgage & no clocks & \IdFmNoClocksAuc{} & \RvFmNoClocksUnitOv{} & \RvFmNoClocksUnitDj{} & \RvFmNoClocksUnitDelta{} & \RvFmNoClocksTimeDelta{} & \RvFmNoClocksGapDelta{} \\
Drives & full & \IdBbFullAuc{} & \RvBbFullUnitOv{} & \RvBbFullUnitDj{} & \RvBbFullUnitDelta{} & \RvBbFullTimeDelta{} & \RvBbFullGapDelta{} \\
Drives & no clocks & \IdBbNoClocksAuc{} & \RvBbNoClocksUnitOv{} & \RvBbNoClocksUnitDj{} & \RvBbNoClocksUnitDelta{} & \RvBbNoClocksTimeDelta{} & \RvBbNoClocksGapDelta{} \\
Clickstreams & full & \IdOuladFullAuc{} & \RvOuladFullUnitOv{} & \RvOuladFullUnitDj{} & \RvOuladFullUnitDelta{} & \RvOuladFullTimeDelta{} & \RvOuladFullGapDelta{} \\
Clickstreams & totals & \IdOuladTotalsAuc{} & \RvOuladTotalsUnitOv{} & \RvOuladTotalsUnitDj{} & \RvOuladTotalsUnitDelta{} & \RvOuladTotalsTimeDelta{} & \RvOuladTotalsGapDelta{} \\
Vessels & full & \IdMarFullAuc{} & \RvMarFullUnitOv{} & \RvMarFullUnitDj{} & \RvMarFullUnitDelta{} & \RvMarFullTimeDelta{} & \RvMarFullGapDelta{} \\
Vessels & no location & \IdMarNoLocAuc{} & \RvMarNoLocUnitOv{} & \RvMarNoLocUnitDj{} & \RvMarNoLocUnitDelta{} & \RvMarNoLocTimeDelta{} & \RvMarNoLocGapDelta{} \\
\bottomrule
\end{tabular}
\caption{Controlled case-control experiment. Each $\Delta$ compares
two fits on the same test rows with the same per-class training
counts, replacing some other-unit rows with test-unit rows in the
overlap arm. Overlap and Disjoint are the unit contrast's mean AUCs.
ICU and
clickstream time is counted from each unit's start, so their calendar
and embargo columns hold out late-in-record rows rather than later
history. Mean-fill and model-fill event fits are fit on training rows;
the associated identity audits use panel-wide fits. Intervals are in the text and
released results.}
\label{tab:bridge}
\end{table}

\textbf{Longer carry-forward increases the measured overlap gap.} In the ICU, the
five variants retain the same six-hour measurement counts and differ
only in how missing values are filled. The patient-overlap gap is
\RvIcuRawUnitDelta{} on raw values and \RvIcuMeanTrainUnitDelta{} with mean
fill, rises to \RvIcuLocfSixUnitDelta{} when values are carried
forward for up to six hours, and reaches \RvIcuLocfUnitDelta{}
[\RvIcuLocfUnitLo, \RvIcuLocfUnitHi] when they are carried forward
without limit. The paired difference between unlimited carry-forward
and raw values is \RvPairIcuLocfRawUnitDelta{}
[\RvPairIcuLocfRawUnitLo, \RvPairIcuLocfRawUnitHi]. Identity AUC shows the same increase with carry-forward duration
(raw \RvIdRawAuc{}, mean fill \RvIdMeanAuc{}, six-hour
\RvIdLocfSixAuc{}, unlimited \RvIdLocfAuc{}); raw and mean fill are
close on both.

\textbf{A model-based fill falls between raw values and unlimited
carry-forward.} The
fifth variant fills each missing value with a regression prediction
from the hour's observed values, each variable's last earlier value
and the counts (chained ridge regressions, fit on training rows only).
Its overlap gap is \RvIcuModelTrainUnitDelta{}
[\RvIcuModelTrainUnitLo, \RvIcuModelTrainUnitHi],
\RvPairIcuModelTrainRawUnitDelta{} above raw values and level with
six-hour carry-forward (difference
\RvPairIcuModelTrainLocfSixUnitDelta{}
[\RvPairIcuModelTrainLocfSixUnitLo,
\RvPairIcuModelTrainLocfSixUnitHi]). Its identity AUC is
\RvIdModelAuc{}, above raw values and above six-hour carry-forward.
The audit therefore separates the fills that draw on a patient's own
history from those that do not; it does not order the three
history-based fills by the size of their gaps. Filling with training-row means
rather than one panel-wide constant changes the mean-fill $\Delta$ by
at most \RvIcuMeanFillMaxDiff{}.

\textbf{The rows with the lowest identity AUC show no clear overlap
gap.}
Table~\ref{tab:bridge} has \RvSepCells{} rows across five domains,
plotted in Figure~\ref{fig:overview}B. The overlap gap's range
excludes zero in \RvSepHighN{} of them: every ICU variant, mortgages,
drives, and vessels with all features. It includes zero in the other
\RvSepLowN{}, which are also the rows with the lowest identity AUCs:
clickstreams ($\Delta$s \RvOuladFullUnitDelta{} [\RvOuladFullUnitLo,
\RvOuladFullUnitHi] and \RvOuladTotalsUnitDelta{}
[\RvOuladTotalsUnitLo, \RvOuladTotalsUnitHi]) and vessels without
position (\RvMarNoLocUnitDelta{} [\RvMarNoLocUnitLo,
\RvMarNoLocUnitHi]). The rows are not independent, since five are ICU
variants and each other domain contributes two, so this pattern
describes these datasets and is not a threshold.

The score does not give the size of a gap. Vessels with all features
read \IdMarFullAuc{} and show \RvMarFullUnitDelta{} [\RvMarFullUnitLo,
\RvMarFullUnitHi]; a casualty vessel has only a few windows, so little
of a test vessel reaches training, which likely limits the gap. ICU six-hour carry-forward reads
\RvIdLocfSixAuc{} and shows \RvIcuLocfSixUnitDelta{}.

\textbf{Substantial identity discrimination remains within hospital.} Matching impostors on hospital
lowers identity AUC from \RvIdLocfAuc{} to \RvIdSiteLocfAuc{} under
carry-forward and from \RvIdRawAuc{} to \RvIdSiteRawAuc{} on raw
values, and a hospital-holdout arm shows no clear imputation effect
(Appendix~\ref{app:extra-arms}).

\begin{table}[H]
\centering
\small
\begin{tabular}{lcccc}
\toprule
Fill & Identity AUC & Overlap & Disjoint & $\Delta$ unit [range] \\
\midrule
carry-forward & \RvIdIcuBLocfAuc{} & \RvIcuBLocfUnitOv{} & \RvIcuBLocfUnitDj{} & \RvIcuBLocfUnitDelta{} [\RvIcuBLocfUnitLo, \RvIcuBLocfUnitHi] \\
carry-forward 6 h & \RvIdIcuBLocfSixAuc{} & \RvIcuBLocfSixUnitOv{} & \RvIcuBLocfSixUnitDj{} & \RvIcuBLocfSixUnitDelta{} [\RvIcuBLocfSixUnitLo, \RvIcuBLocfSixUnitHi] \\
raw & \RvIdIcuBRawAuc{} & \RvIcuBRawUnitOv{} & \RvIcuBRawUnitDj{} & \RvIcuBRawUnitDelta{} [\RvIcuBRawUnitLo, \RvIcuBRawUnitHi] \\
mean fill & \RvIdIcuBMeanAuc{} & \RvIcuBMeanTrainUnitOv{} & \RvIcuBMeanTrainUnitDj{} & \RvIcuBMeanTrainUnitDelta{} [\RvIcuBMeanTrainUnitLo, \RvIcuBMeanTrainUnitHi] \\
\bottomrule
\end{tabular}
\caption{Replication on PhysioNet 2012, event in-hospital death. Same
design as Table~\ref{tab:bridge}: fixed test rows, matched per-class
training counts, 200 replicates. Every fill keeps the six-hour
measurement counts.}
\label{tab:icu2012}
\end{table}

\textbf{The result replicates on a second ICU dataset.} The
PhysioNet 2012 challenge data \citep{silva2012predicting} contain
12{,}000 ICU stays from MIMIC-II. Using hourly rows over the first
48 hours, in-hospital death as the event, and the
same fills, counts and controls (Table~\ref{tab:icu2012}) gives
the same unit-overlap pattern. Identity
AUC and overlap gap rise in the same steps. Unlimited carry-forward
minus raw values is \RvPairIcuBLocfRawUnitDelta{}
[\RvPairIcuBLocfRawUnitLo, \RvPairIcuBLocfRawUnitHi] on the gap,
six-hour minus raw is \RvPairIcuBLocfSixRawUnitDelta{}
[\RvPairIcuBLocfSixRawUnitLo, \RvPairIcuBLocfSixRawUnitHi], and mean
fill minus raw is \RvPairIcuBMeanTrainRawUnitDelta{}
[\RvPairIcuBMeanTrainRawUnitLo, \RvPairIcuBMeanTrainRawUnitHi].
Carry-forward leads raw values by \RvIcuBLocfRawOvLead{} AUC with
shared patients and by \RvIcuBLocfRawDjLead{} held out. One of the two
PhysioNet 2019 hospital systems is the source of MIMIC-II, so this is
a second dataset and a second outcome, not an independent hospital.

\subsection{Holding Out the Future Does Not Hold Out the Unit}
\label{sec:calendar-arms}

A temporal holdout need not exclude units seen in training.
Whether retaining those units is appropriate depends on deployment. At the 80th-percentile cut, \BbStraddle{} of
drives, \FmStraddle{} of loans, \OuladStraddle{} of learners and
\IcuStraddle{} of stays have rows on both sides. In the ICU and
clickstream panels time is counted from each unit's own start, so no
absolute-calendar split can be built from these time columns. On
mortgages and drives, the overlap gap survives the
calendar split in every cell: mortgages \RvFmFullTimeDelta{}
[\RvFmFullTimeLo, \RvFmFullTimeHi] and \RvFmNoClocksTimeDelta{}
[\RvFmNoClocksTimeLo, \RvFmNoClocksTimeHi], drives
\RvBbFullTimeDelta{} [\RvBbFullTimeLo, \RvBbFullTimeHi] and
\RvBbNoClocksTimeDelta{} [\RvBbNoClocksTimeLo, \RvBbNoClocksTimeHi].
An embargo between training and test helps unevenly. On mortgages
the gap stays positive (\RvFmFullGapDelta{} [\RvFmFullGapLo,
\RvFmFullGapHi]); on drives the interval covers zero
(\RvBbFullGapDelta{} [\RvBbFullGapLo, \RvBbFullGapHi]). On vessels both ranges include zero
(calendar \RvMarFullTimeDelta{} [\RvMarFullTimeLo, \RvMarFullTimeHi],
embargo \RvMarFullGapDelta{} [\RvMarFullGapLo, \RvMarFullGapHi]).
A calendar split and a unit split are separate
controls, and applying one does not provide the other.

\section{Fixed-Horizon Forecasting}
\label{sec:forecast}

I also test whether a model can predict recorded sepsis onset
within a fixed time horizon using ICU records. Onset is six hours after the first positive
challenge label. At hour $t$, the target is recorded onset within the
next \RvFcHorizon{} hours. Rows at or after onset and \RvFcCensored{}
stays positive from their first recorded hour are excluded; the latter
have no identifiable onset from the shifted-label transition.
Raw values use the current hour; carry-forward and measurement counts
use current and past hours. Reported mean-fill fits estimate constants
from training rows.

\subsection{Random Patient-Hour Overlap}

Each of 200 replicates retains all \RvFcPosUnits{} eligible septic
stays, samples 4{,}000 never-septic stays, and includes every positive
hour and up to 20 negative hours per stay. Some septic stays have no
eligible negative hour. Negatives mean no recorded onset in the
horizon; no additional follow-up requirement is imposed near the end
of a record. The unit contrast randomly divides sampled test-patient
hours: overlap training can therefore include later hours and labels
from those patients. Section~\ref{sec:chrono} compares this design
with training restricted to earlier, resolved patient hours.

\begin{table}[H]
\centering
\small
\begin{tabular}{lccccc}
\toprule
Imputation & Identity AUC & Overlap & Disjoint & $\Delta$ & 95\% range \\
\midrule
carry-forward & \RvIdLocfAuc{} & \RvFcLocfUnitOv{} & \RvFcLocfUnitDj{} & \RvFcLocfUnitDelta{} & [\RvFcLocfUnitLo, \RvFcLocfUnitHi] \\
carry-forward 6 h & \RvIdLocfSixAuc{} & \RvFcLocfSixUnitOv{} & \RvFcLocfSixUnitDj{} & \RvFcLocfSixUnitDelta{} & [\RvFcLocfSixUnitLo, \RvFcLocfSixUnitHi] \\
model fill & \RvIdModelAuc{} & \RvFcModelTrainUnitOv{} & \RvFcModelTrainUnitDj{} & \RvFcModelTrainUnitDelta{} & [\RvFcModelTrainUnitLo, \RvFcModelTrainUnitHi] \\
raw & \RvIdRawAuc{} & \RvFcRawUnitOv{} & \RvFcRawUnitDj{} & \RvFcRawUnitDelta{} & [\RvFcRawUnitLo, \RvFcRawUnitHi] \\
mean fill & \RvIdMeanAuc{} & \RvFcMeanTrainUnitOv{} & \RvFcMeanTrainUnitDj{} & \RvFcMeanTrainUnitDelta{} & [\RvFcMeanTrainUnitLo, \RvFcMeanTrainUnitHi] \\
\bottomrule
\end{tabular}
\caption{ICU sepsis forecasting, onset within \RvFcHorizon{} hours.
Approximately half of each test patient's sampled hours enter the
overlap training pool. Both fits score the same test hours with
matched per-class training counts. Ranges are empirical replicate
percentiles. Identity AUC for mean and model fill uses panel-wide
fits, whereas forecasting fits them on training rows.}
\label{tab:forecast}
\end{table}

Patient overlap raises AUC by \RvFcLocfUnitDelta{} under carry-forward
and \RvFcRawUnitDelta{} on raw values, a paired difference of
\RvPairFcLocfRawUnitDelta{} [\RvPairFcLocfRawUnitLo,
\RvPairFcLocfRawUnitHi] (Table~\ref{tab:forecast}). The gap is larger
than in case-control classification. Disjoint AUCs remain different
(\RvFcLocfUnitDj{} versus \RvFcRawUnitDj{}): carry-forward also improves
this measured disjoint performance. Identity AUC and the overlap gap
both increase with carry-forward duration, while raw and mean fill
are close. This association is observed in one domain and one horizon,
not a validated rule for predicting gap magnitudes.

\textbf{A shared-patient comparison can pick the wrong fill.} With
shared patients, unlimited carry-forward leads the model fill by
\RvFcLocfModelOvLead{} AUC; with patients held out the two are level
(\RvFcLocfModelDjLead{}). Six-hour carry-forward is level with the
model fill when patients are shared (\RvFcModelLocfSixOvLead{} for the
model fill) and behind it when they are held out
(\RvFcModelLocfSixDjLead{}); the difference in their overlap gaps is
\RvPairFcModelTrainLocfSixUnitDelta{}
[\RvPairFcModelTrainLocfSixUnitLo, \RvPairFcModelTrainLocfSixUnitHi].

\subsection{Within-Patient Chronological Control}
\label{sec:chrono}

This comparison requires complete follow-up for negative labels and
restricts training to a test patient's earlier hours whose labels
have resolved by the prediction cut. For each test patient the cut
is the ceiling of the median sampled hour. Test hours are at or
after the cut; training hours must end their prediction horizon by
the cut. Negative hours without a full \RvFcHorizon{}-hour follow-up
are excluded. Both arms score identical test rows with matched
per-class training counts.

\begin{table}[H]
\centering
\small
\begin{tabular}{lcc}
\toprule
Imputation & Random overlap $\Delta$ [range] & Chronological overlap $\Delta$ [range] \\
\midrule
raw & \RvChRawUnitDelta{} [\RvChRawUnitLo, \RvChRawUnitHi] & \RvChRawChronoDelta{} [\RvChRawChronoLo, \RvChRawChronoHi] \\
mean fill & \RvChMeanTrainUnitDelta{} [\RvChMeanTrainUnitLo, \RvChMeanTrainUnitHi] & \RvChMeanTrainChronoDelta{} [\RvChMeanTrainChronoLo, \RvChMeanTrainChronoHi] \\
carry-forward 6 h & \RvChLocfSixUnitDelta{} [\RvChLocfSixUnitLo, \RvChLocfSixUnitHi] & \RvChLocfSixChronoDelta{} [\RvChLocfSixChronoLo, \RvChLocfSixChronoHi] \\
carry-forward & \RvChLocfUnitDelta{} [\RvChLocfUnitLo, \RvChLocfUnitHi] & \RvChLocfChronoDelta{} [\RvChLocfChronoLo, \RvChLocfChronoHi] \\
\bottomrule
\end{tabular}
\caption{Sepsis forecasting with complete follow-up. Each column
compares overlap and disjoint training on identical test rows with
matched training counts. Test rows differ between columns because
the chronological design uses only hours at or after each patient's
cut. Ranges are empirical percentiles over 200 replicates.}
\label{tab:chrono}
\end{table}

Requiring complete follow-up preserves the random-overlap effect:
carry-forward minus raw is \RvPairChLocfRawUnitDelta{}
[\RvPairChLocfRawUnitLo, \RvPairChLocfRawUnitHi] on the gap.
The chronological comparison shows no positive mean gap for any fill
(Table~\ref{tab:chrono}). Earlier hours contribute training data for
\RvChContribPatients{} of \RvChTestPatients{} test patients. All
admissible hours are negative, so this control changes both the
information and the amount of training data supplied by test
patients. High identity AUC accompanies inflation under random
patient-hour overlap, but not under this chronological protocol.

\subsection{The Clinical Values Add Little Beyond the Clock and
Measurement Counts}
\label{sec:clock}

On the complete-follow-up cohort, hour of stay plus measurement
counts, with no clinical values, scores \RvChCountsTimeUnitDj{} on
held-out patients. Adding the raw clinical values gives
\RvChRawTimeUnitDj{}, a difference of \RvChLeadRawTimeCountsTimeDj{}
[\RvChLeadRawTimeCountsTimeDjLo, \RvChLeadRawTimeCountsTimeDjHi], so
the clinical values add a small increment to what the clock and the
measurement pattern already give. Hour of stay alone scores
\RvChTimeOnlyUnitDj{}. The clinical-value variants of
Table~\ref{tab:chrono} omit hour of stay, and without it raw values
with counts score \RvChRawUnitDj{}, below hour of stay alone
(difference \RvChLeadTimeOnlyRawDj{} [\RvChLeadTimeOnlyRawDjLo,
\RvChLeadTimeOnlyRawDjHi]), and unlimited carry-forward of every
value scores \RvChLocfUnitDj{}, below hour of stay plus counts
(difference \RvChLeadCountsTimeLocfDj{} [\RvChLeadCountsTimeLocfDjLo,
\RvChLeadCountsTimeLocfDjHi]). Claims that
physiological measurements add predictive value should be evaluated
against these baselines on the same target, cohort and split.

The random-split exaggeration, the chronological control and the
hour-of-stay baseline each hold within each of the two PhysioNet 2019
hospital systems analyzed separately (Table~\ref{tab:sites}).

\begin{table}[H]
\centering
\footnotesize
\setlength{\tabcolsep}{3pt}
\begin{tabular}{llcccc}
\toprule
Hospital & Fill & Overlap & Disjoint & Random-split gap [range] & Chronological gap [range] \\
\midrule
A & carry-forward & \RvStALocfOv{} & \RvStALocfDj{} & \RvStALocfDelta{} [\RvStALocfLo, \RvStALocfHi] & \RvStALocfChDelta{} [\RvStALocfChLo, \RvStALocfChHi] \\
A & carry-forward 6 h & \RvStALocfSixOv{} & \RvStALocfSixDj{} & \RvStALocfSixDelta{} [\RvStALocfSixLo, \RvStALocfSixHi] & \RvStALocfSixChDelta{} [\RvStALocfSixChLo, \RvStALocfSixChHi] \\
A & raw & \RvStARawOv{} & \RvStARawDj{} & \RvStARawDelta{} [\RvStARawLo, \RvStARawHi] & \RvStARawChDelta{} [\RvStARawChLo, \RvStARawChHi] \\
B & carry-forward & \RvStBLocfOv{} & \RvStBLocfDj{} & \RvStBLocfDelta{} [\RvStBLocfLo, \RvStBLocfHi] & \RvStBLocfChDelta{} [\RvStBLocfChLo, \RvStBLocfChHi] \\
B & carry-forward 6 h & \RvStBLocfSixOv{} & \RvStBLocfSixDj{} & \RvStBLocfSixDelta{} [\RvStBLocfSixLo, \RvStBLocfSixHi] & \RvStBLocfSixChDelta{} [\RvStBLocfSixChLo, \RvStBLocfSixChHi] \\
B & raw & \RvStBRawOv{} & \RvStBRawDj{} & \RvStBRawDelta{} [\RvStBRawLo, \RvStBRawHi] & \RvStBRawChDelta{} [\RvStBRawChLo, \RvStBRawChHi] \\
\bottomrule
\end{tabular}
\caption{Sepsis forecasting within each PhysioNet 2019 hospital
system, complete-follow-up cohort, 200 replicates each. The
carry-forward minus raw difference in random-split gaps is
\RvStAPairLocfRawDelta{} [\RvStAPairLocfRawLo, \RvStAPairLocfRawHi] in
hospital A and \RvStBPairLocfRawDelta{} [\RvStBPairLocfRawLo,
\RvStBPairLocfRawHi] in hospital B. Hour of stay alone scores
\RvStATimeOnlyDj{} and \RvStBTimeOnlyDj{} on held-out patients,
\RvStALeadTimeOnlyRaw{} [\RvStALeadTimeOnlyRawLo,
\RvStALeadTimeOnlyRawHi] and \RvStBLeadTimeOnlyRaw{}
[\RvStBLeadTimeOnlyRawLo, \RvStBLeadTimeOnlyRawHi] above raw values.}
\label{tab:sites}
\end{table}

\section{Using the Audit}
\label{sec:using}

Choose the evaluation split from the intended use before running the
audit. Test on unseen units when deployment concerns new units;
for known units, restrict training to information available before
prediction. Apply these requirements regardless of the audit score.
Then use the audits to locate features and preprocessing choices
for further checks (Table~\ref{tab:using}).

\textbf{Scope and interpretation.} High audit AUC establishes that the
tested representation contains recoverable unit identity or temporal
ordering; it does not by itself establish inflated event-prediction
performance. Such information can support a misleading evaluation when
it is associated with the prediction target and the evaluation
protocol permits its use in ways unavailable in deployment. My
experiments illustrate why recoverability alone is insufficient. Clickstream records identify the
student above chance (identity AUC \IdOuladFullAuc{}) and show no
clear positive overlap gap (\RvOuladFullUnitDelta{} [\RvOuladFullUnitLo,
\RvOuladFullUnitHi]). Carry-forward ICU records identify the patient
almost perfectly (\RvIdLocfAuc{}), and the gap they show under random
patient-hour overlap is absent when training uses only a patient's
earlier, resolved hours (\RvChLocfChronoDelta{} [\RvChLocfChronoLo,
\RvChLocfChronoHi]). For known-unit forecasting, shared identity can
be appropriate when training uses only information available before
prediction. The audits therefore identify features and preprocessing
choices for follow-up comparisons under deployment-aligned splits;
they do not provide a mapping from audit AUC to downstream inflation.

\begin{table}[H]
\centering
\small
\begin{tabular}{p{0.30\linewidth}p{0.62\linewidth}}
\toprule
Audit result & Follow-up comparison \\
\midrule
Identity discrimination increases under preprocessing &
Compare preprocessing choices under the deployment split, using the
same test rows and matched training counts. The tested random-overlap
protocols exaggerate the gains of carry-forward imputation. \\
Timing discrimination falls after feature removal &
Compare the event model with a baseline using those features alone,
provided they are available at prediction time. \\
Timing discrimination survives feature removal &
Compare against a time-only baseline on the same target and test
rows; information about temporal position remains in the features. \\
Scores change with temporal offset &
Take held-out data as far past training as the model will be used:
drive records a month apart are nearly indistinguishable (temporal AUC
\BbOffShortAuc{}) and clearly different after three to six months
(\BbOffLongAuc{}). Waiting does not remove identity; drive identity AUC
is \RvIdOffLongBbFullAuc{} at that gap. \\
Either AUC is near 0.5 &
This probe detected little signal at these offsets. Retain the
evaluation split required by deployment. \\
\bottomrule
\end{tabular}
\caption{Using the audit to choose follow-up comparisons. Audit AUC
measures discrimination on the pairing task; the downstream
comparison measures its consequence for event prediction.}
\label{tab:using}
\end{table}

For example, a sepsis model for new patients requires held-out
patients under every preprocessing choice. Carry-forward's higher
identity AUC motivates checking whether a shared-patient comparison
overstates its benefit. On the complete-follow-up cohort, its
advantage over raw values is \RvChLeadLocfRawOv{} with random overlap
and \RvChLeadLocfRawDj{} with held-out patients. Report the latter
comparison alongside the hour-of-stay baseline from the same cohort
(Section~\ref{sec:clock}).

\section{Related Work}
\label{sec:related}

Both tests apply negative-control reasoning
\citep{lipsitch2010negative}, and each has established predecessors.
For identity, these are biometric verification
\citep{jain2004introduction,odinaka2012ecg} and privacy studies of how
unique individual records are
\citep{demontjoye2013unique,rocher2019estimating}, including
re-identification of patients from laboratory orders alone
\citep{johnson2019reidentification}. For time, the precedents are the self-matching of the
case-crossover design and its time-trend corrections
\citep{maclure1991casecrossover,suissa1995casetimecontrol}, placebo
tests \citep{abadie2015comparative}, and classifier two-sample tests
\citep{lopezpaz2017revisiting,clemencon2009auc}. Both audits are gradient-boosted classifiers on
matched pairs of windows from one panel: the identity test matches
windows on the declared time coordinate and offset, and the temporal
test pairs windows within units. Temporal-order pretext tasks
\citep{misra2016shuffle,wei2018arrow} pose the same ordering question
to learn representations; here it measures a dataset.

Leakage taxonomies \citep{kaufman2012leakage,kapoor2023leakage},
reporting standards \citep{kapoor2024reforms}, meta-reviews of
evaluation failure \citep{liao2021learning}, and a catalog of
forecasting pitfalls \citep{hewamalage2023forecast} describe how evaluations can fail and how to check them. My audits
complement that work by measuring identity and temporal information
in the data before an event-prediction model is fit. Work on dataset
documentation calls for measured, comparable properties of datasets
\citep{mitchell2022measuring}; the two AUCs are such properties, with a
known null expectation. Large-scale leakage measurement compares leaky and clean
evaluation of trained models \citep{roth2026leakage}, which requires classification labels and a task. My audits
require event status to select event-free units, but no forecasting
target.
Where unit identity crossing a split inflated published results
\citep{wen2020convolutional} or, in controlled experiments,
drastically inflated performance \citep{rosenblatt2024leakage}, an
identity audit of the kind used here could have been run on the data
before any of those models were fit. Recognition can also act at the
level of a group: \citet{zech2018variable} show that a chest
radiograph model identifies the hospital system that produced an
image, and that differences in disease prevalence between hospitals
then confound its predictions. My identity test concerns the
individual unit; the most-similar-unit pairs of
Table~\ref{tab:nearest} and the hospital-matched pairs and cohort
holdouts of Appendix~\ref{app:extra-arms} address the group level.
Appendix~\ref{sec:relatedfull} compares the closest methods.

\section{Limitations}
\label{sec:limitations}

Audit AUC depends on the probe, feature set, pairing population and
offset range. The six-domain results describe these declared
measurements; different task designs prevent interpreting identity
and temporal AUCs as a common information scale. Matching controls
the panel's time coordinate, not every shared environmental factor:
hours since admission do not match admission dates. Hospital-matched
results address one such factor (Appendix~\ref{app:extra-arms}).

The imputation effect holds on two ICU datasets that share a source
hospital. Forecasting covers one dataset and one horizon; sampled
hours do not represent deployment prevalence. Replicate ranges
describe variation conditional on the available pools. Unit overlap
can be appropriate for known-unit deployment when training uses only
information available before prediction, as the chronological control
illustrates. The audits do not detect leakage introduced by resampling
before splitting \citep{vandewiele2021overly} or model selection on
evaluation data \citep{neunhoeffer2019crossvalidation}.

\section{Conclusion}
\label{sec:conclusion}

Ordinary records reveal which unit they came from and when, and two
simple audits measure how much before any event-prediction model is
trained. Across two ICU datasets, carry-forward imputation makes
patients far easier to recognize and exaggerates its own measured
benefit when a patient's hours are split between training and test. On
the sepsis task, a within-patient chronological control shows no
positive mean gap, so recognizability does not become inflation under
every protocol.
With complete follow-up on the same task, hour of stay and
measurement counts, with no clinical value, score
\RvChCountsTimeUnitDj{} on held-out patients, and the raw clinical
values add \RvChLeadRawTimeCountsTimeDj{}. On drives and mortgages, a
calendar split leaves most units on both sides and the overlap gap in
place. The audits tell a practitioner which features and
preprocessing choices to check, and the released code runs both the
audits and the checks.

\section*{Reproducibility Statement}

The supplementary material contains the complete harness (an
installable package with invariant tests for the forced-null, split,
placebo, and controlled-overlap constructions), panel adapters, declarations and stored results for the six-domain
harness measurements, the controlled overlap and forecasting
experiments with their stored results, the PhysioNet 2012 adapter,
and the scripts that build the ICU and drive
panels from the public raw downloads. The vessel adapter reads
per-window summaries published with the linked maritime corpus, whose
computation from raw tracks is not included.
Appendix~\ref{app:harness} and the supplementary README state these
boundaries.
The supplementary material also includes the manuscript source, in
which every reported AUC, gap and interval is generated from the
stored result files.
Sections~\ref{sec:definition} and~\ref{sec:eligibility} specify the
construction and its preconditions;
Appendix~\ref{app:validation} validates the interval machinery.
The pipeline is seeded throughout, and a full regeneration of the 30
temporal, identity and placebo measurement files reproduces the stored
per-seed AUCs and interval bounds to the fourth decimal.

Public-data runs require the corresponding source downloads and
panel preparation. The package also includes a self-contained
demonstration, \texttt{aa-floor-demo}, that needs no download and no
credentials. It synthesizes a panel with signal planted on known
columns, runs both tests on it through the same entry points
used for the six-domain audit measurements, and checks each tier
against what was planted: each test must read its own axis above
the null, read the other test's axis at the null, and read pure
noise at the null. It reports whether each feature tier meets its expected result and exits non-zero if
any tier misses. Appendix~\ref{app:harness} gives the construction.

I also release \texttt{whowhen} version 0.3.1
(\url{https://github.com/mojolearn/whowhen}; \citealp{hendel2026whowhen}), a Python library that
runs identity and timing audits on panel tables. The user names the unit
and time columns; the report includes re-identification, time-order and
timing-placebo scores. When event labels permit, it also provides
corresponding evaluation checks for overlap, chronological and calendar
splits, and comparison with time alone. The library fits its classifiers
with mojolearn \citep{hendel2026mojolearn}. For the release used here,
the numerical outputs of the turbine report agreed exactly between an
Apple M4 and an NVIDIA RTX 4090; execution metadata is excluded from
that comparison. Appendix~\ref{app:library} reports the library's audit
values on the six domains and specifies its protocol. Every other number
in this paper comes from the supplementary research harness, which uses
a different classifier.

\section*{AI Use Statement}

Large language models were used to retrieve and discover related
literature, support research ideation and hypothesis refinement,
assist with methodology and experiment design, and aid software
implementation (including the code that synthesizes the released
demonstration's panel from a random seed; no model produced any
data), mathematical formulation and proof assistance for
Proposition~\ref{prop:forced-null} and its proof in
Appendix~\ref{app:proof}, interpretation of results, and manuscript
drafting and editing, including drafting sections and polishing their
writing. Empirical values are outputs of executed experiments, with
the synthetic demonstration distinguished from the public-data
measurements. Harness values pass through generated macros and
transcription audits. These audits check
consistency with stored results, not the validity or provenance of
the experiments themselves. The research questions and study design
were chosen, the experiments executed and audited, the proof checked,
and the reported results and citations verified by the human
authorship of this paper, which takes responsibility for its
contents.

\section*{Broader Impact}
The audits help practitioners find evaluation results that
overstate how a model will perform on new patients, drives or loans,
which matters most where predictions inform clinical or financial
decisions. All measurements use public, de-identified datasets under
their published terms. The identity test asks only whether two windows
\emph{within} a dataset share an anonymous unit label; no window is
linked to an external identity, and results are reported as aggregate
AUCs. The same measurement shows that de-identified operational
records remain linkable to one another within a dataset, in the ICU
from measurement counts alone. That is useful to data stewards
assessing a release, and it could also inform an attempt to link
records across releases of the same data; I report aggregate
discrimination only and release no linkage of individual records. A
near-chance audit score does not certify an evaluation, and reading it
as a certificate could give false assurance.

\section*{Competing Interests}
The author declares no competing interests.

\vskip 0.2in
\bibliography{refs}

\appendix

\section{Datasets}
\label{app:domains}

\begin{table}[H]
\centering
\footnotesize
\setlength{\tabcolsep}{4pt}
\begin{tabular}{llllc}
\toprule
Domain & Source & Units & Sampling & Offsets \\
\midrule
ICU & PhysioNet 2019 \citep{reyna2020sepsis} & \IcuRetPanel{} stays & hourly & 6--24 h \\
ICU (2012) & PhysioNet 2012 \citep{silva2012predicting} & 12{,}000 stays & hourly & 6--24 h \\
Drives & Backblaze \citep{backblaze2026data} & \BbRetPanel{} drives & daily & 30--90 d \\
Mortgage & Freddie Mac \citep{freddiemac2026loanlevel} & \FmRetPanel{} loans & monthly & 6--18 mo \\
Clickstreams & OULAD \citep{kuzilek2017oulad} & \OuladRetPanel{} registrations & active days & 14--60 d \\
Wind & \citet{plumley2023kelmarsh,plumley2023penmanshiel} & \WindRetPanel{} turbines & daily & 14--60 d \\
Vessels & Linked AIS corpus \citep{hendel2026corpus} & \MarRetPanel{} vessels & weekly windows & 14--60 d \\
\bottomrule
\end{tabular}
\caption{The six panels and the second ICU dataset used for replication. Each is mapped to the panel format by an
adapter, and a declaration fixes its offsets, feature tiers and clock
columns.}
\label{tab:datasets}
\end{table}

Table~\ref{tab:datasets} lists the panels. The drive panel's clock columns are power-on hours and total writes.
The mortgage clocks are balance, age and remaining term; without them
the remaining timing signal is carried by the quarterly loan-to-value
index. Clickstream windows are measured on days a student is active.
For the overlap experiment the vessel panel adds the corpus's 768
casualty vessels, each with its pre-incident window and earlier windows
of the same vessel. The audited vessel panel is the casualty-free pool of a public corpus that
links casualty investigations to AIS tracks, with one window about
every seven days over two years, summarized into kinematics, history
statistics, AIS reporting counts and position.

\textbf{Drives are identified individually, not by model.} Drive
identity AUC is \IdBbFullAuc{} with clocks and \IdBbNoClocksAuc{}
without, so the wear-state pattern alone is unit-specific.
Restricting every impostor to the anchor's own drive model, so that
model and capacity carry no information, leaves \IdBbCohortFullAuc{}
[\IdBbCohortFullCiLo, \IdBbCohortFullCiHi], so the classifier is not
only recognizing drive models.

\textbf{Vessels are identifiable without position.} Vessel identity
AUC is \IdMarFullAuc{} [\IdMarFullCiLo, \IdMarFullCiHi] on windows
whose temporal AUC is \MarFullAuc{}. Removing latitude and longitude
leaves \IdMarNoLocAuc{}: speed profile and reporting frequency identify
the vessel, and reporting frequency alone gives \IdMarRepAuc{}, the
counterpart of the ICU measurement-count result.

\textbf{Twenty turbines are too few for the identity test.} Five-seed
means range \IdWindFullSeedLo--\IdWindFullSeedHi{} and the pooled
interval [\IdWindFullCiLo, \IdWindFullCiHi] spans more than a third
of the AUC axis, and a single draw (\IdWindPilotAuc{}) is not
representative of its own seeds. Pooled temporal scoring draws power
from thousands of rows, while the identity test's effective sample
size is the number of units.

\textbf{The mortgage rate-only tier reads the null.} The loan's fixed
rate is constant within a unit, and its temporal AUC is \FmRateAuc{}
to four decimals, the value symmetry forces.

\section{Harness}
\label{app:harness}

The released harness measures both quantities, and the placebo, for
any eligible panel from one parquet file and one JSON declaration.
Per-seed values and intervals ship in the supplementary material, and
the numbers in this paper are generated from those files.

\texttt{aa-floor-demo} needs no download. It synthesizes a panel with
a timing signal on one group of columns (a calendar counter and a
trend shared by every unit), an identity signal on another (per-unit
constant offsets plus noise) and pure noise on a third. It then runs
the published entry points unmodified and checks that each test
detects its own signal and reads 0.5 on the other's and on the noise.

\section{Proof of the Forced Null}
\label{app:proof}

\begin{proposition}[Forced null]
\label{prop:forced-null}
Let the test set be $m$ matched pairs from the declared protocol,
pair $k$ contributing a scored item $U_k$ labeled 1 and an item
$V_k$ labeled 0, and let $s$ be any scorer that is a fixed
measurable function of the features and independent of the held-out
labels. If the joint law of $((U_1,V_1),\ldots,(U_m,V_m))$ is
invariant to exchanging $U_k$ with $V_k$ for every $k$ at once, then
the pooled empirical AUC with mid-rank ties satisfies
$\mathbb{E}[\widehat{\mathrm{AUC}}] = 1/2$, for every such $s$ and
every $m$.
\end{proposition}

\begin{proof}
Write the pooled empirical AUC with the mid-rank tie correction as
\begin{equation*}
\widehat{\mathrm{AUC}}
= \frac{1}{m^{2}} \sum_{k=1}^{m} \sum_{l=1}^{m}
\Big[ \mathbf{1}\{s(U_k) > s(V_l)\}
+ \tfrac{1}{2}\mathbf{1}\{s(U_k) = s(V_l)\} \Big].
\end{equation*}
Let $\sigma$ exchange $U_k$ with $V_k$ for every $k$ at once. By
hypothesis $\sigma$ preserves the joint law of the test sample, and
$s$ is held fixed, so $\widehat{\mathrm{AUC}} \circ \sigma$ has the
same distribution as $\widehat{\mathrm{AUC}}$. Relabeling the
summation indices, the $(k,l)$ term of
$\widehat{\mathrm{AUC}} \circ \sigma$ pairs with the $(k,l)$ term of
$\widehat{\mathrm{AUC}}$ to give
$\mathbf{1}\{a > b\} + \mathbf{1}\{b > a\} + \mathbf{1}\{a = b\} = 1$
with $a = s(U_k)$ and $b = s(V_l)$, so
$\widehat{\mathrm{AUC}} \circ \sigma = 1 - \widehat{\mathrm{AUC}}$
pointwise. Taking expectations,
$\mathbb{E}[\widehat{\mathrm{AUC}}]
= \mathbb{E}[1 - \widehat{\mathrm{AUC}}]$, and the claim follows.
\end{proof}

\paragraph{Why the ordered pair and not the marginals.} Equal
marginals do not force the null, because the two items of a pair are
dependent and the within-pair comparisons enter the statistic. The
temporal null is therefore exchangeability of the ordered window
pair, which is strictly stronger than stationarity of $X_{i,t}$ and
is the sense in which the paper uses the word throughout. The
identity null is already in this form, the same-unit and
different-unit continuations of a shared anchor being exchangeable
once calendar and offset are matched.

\paragraph{What the statement fixes.} \emph{Randomness.} The
expectation is over the draw of the test pairs, with the training
units, the fitted scorer, and the scorer's own seed held fixed by
conditioning. The published values additionally average over
pair-draw seeds and the intervals resample units; both sit outside
the proposition and are reported separately.
\emph{Independence.} The scorer must not depend on the held-out
labels. The unit-disjoint split keeps the held-out units out of fitting, and in the identity
test the units are split before any pair is drawn, so no
unit contributes to both fitting and scoring.
\emph{Ties.} Gradient-boosted ensembles produce many ties, so
the mid-rank convention matters. The involution needs the
$\tfrac{1}{2}$ weight for the three indicators to sum to one, and it
is the convention the reported AUCs already use.
\emph{Scope of ``every classifier.''} Every scorer measurable in the
features and independent of the held-out labels, which covers any
model fit on the training units by any procedure. It excludes a
procedure that reads the held-out labels, which is the failure the
design prevents. \emph{Finite sample.} The conclusion
concerns the expected empirical AUC at the realized $m$ and not a
population limit, so the bootstrap width and the seed spread quoted
throughout describe spread around a mean that the null fixes at
exactly one half.

\paragraph{Where the released construction departs from the
idealization.} The proposition assumes every anchor yields a matched
pair. In the identity test the units are split before any
pair is drawn and the impostor is drawn from the anchor's own side,
so the split discards nothing; the one residue is an anchor whose
matched time offers no impostor on its side, which emits its
same-unit row unmatched. The proposition applies to the matched subsample under its stated
joint-invariance assumption; the unmatched residue requires a separate
qualification. Class imbalance alone does not establish AUC bias.
The harness records the
per-side class counts in every results file, and
Table~\ref{tab:identity-balance} replays the published seeds and
reports them: the test sides are balanced to within the residue,
which is zero on every large panel, one or two rows under
cohort-matched impostors, and at most fifteen on the twenty-turbine
wind panel, where a matched time sometimes offers no impostor among
four test units. The temporal
test has no analogous residue, since both windows of a pair
come from the same unit by construction. Neither test
assumes the null is true. The proposition gives the expected reading
if it were true, which is why AUC above 0.5 can be read as signal in
the data.

\begin{table}[H]
\centering
\scriptsize
\input{identity_balance.tex}
\caption{Identity matching diagnostics, ranges across the five
published seeds, replayed from the stored seeds without refitting.
Train same is the number of same-unit training pairs; test same and
test different are the two test classes; test unmatched counts
same-unit test rows whose matched time offered no impostor on the
test side. Per-seed counts ship in
\texttt{results/identity\_balance.json} and in each results file.}
\label{tab:identity-balance}
\end{table}

\section{Interval Validation and Additional Controls}
\label{app:validation}

Checks on the measurement machinery. \textbf{Reading the pooled
interval.} A single-seed interval conditions on that seed's pairs,
which is how a five-seed mean can fall outside it. The pooled
interval describes the spread of one measurement while the quoted
point averages five, so it describes a different target from a
confidence interval for the five-seed mean. No general coverage
guarantee is claimed. \textbf{Coverage.} Identity intervals come from
a dyadic cluster bootstrap that resamples units over both pair roles.
It was compared against a 200-repetition full-regeneration bootstrap on the
two headline unsaturated cells, which resamples units, redraws pairs,
refits, and rescores, and so includes every source of variation. Regeneration
intervals [\RegenOuladCiLo, \RegenOuladCiHi] (clickstreams) and
[\RegenIcuRawCiLo, \RegenIcuRawCiHi] (ICU values only) sit inside the reported
pooled intervals [\IdOuladFullCiLo, \IdOuladFullCiHi] and
[\IdIcuRawFullCiLo, \IdIcuRawFullCiHi], with regeneration means within
\RegenMeanGapMax{} of the five-seed means. Pooling widens the dyadic
interval relative to its primary-seed version ([\IdOuladFullPrimaryLo,
\IdOuladFullPrimaryHi] and [\IdIcuRawFullPrimaryLo,
\IdIcuRawFullPrimaryHi]), so the pooled intervals are wider than the regeneration intervals
in these two checked cells. This comparison does not establish nominal
coverage or conservativeness on other panels. \textbf{Cohort matching.} Drive
impostors restricted to the anchor's model leave identity AUC at
\IdBbCohortFullAuc{}, and ICU impostors matched on hospital are in
Appendix~\ref{app:extra-arms}; the equivalent test has not been run
for clickstreams, where course cohort remains a possible explanation
for part of that number.
\textbf{Retention.} The pairing rule
narrows no domain by more than about a quarter, and eligible units
exceed 90\% of ordinary units in every large panel except vessels,
whose weekly windows leave \MarRetEligible{} of \MarRetOrdinary{}
with two windows 14--60 days apart. Retention ships with every
result.

\section{Scorer Sensitivity}
\label{app:scorer}

Full-tier temporal AUCs under the harness tree and under a linear
classifier (logistic regression, median-imputed, standardized; five seeds
each). Mortgage \FmFullAuc{} vs \LinFmFullAuc{}, ICU \IcuFullAuc{} vs
\LinIcuFullAuc{}, drives \BbFullAuc{} vs \LinBbFullAuc{}, wind
\WindFullAuc{} vs \LinWindFullAuc{}, clickstreams \OuladFullAuc{} vs
\LinOuladFullAuc{}, vessels \MarFullAuc{} vs \LinMarFullAuc{}. The mortgage rate-only rows agree across classifiers to the
fourth decimal. All shifts are downward or within noise, consistent
with each fitted AUC being at most its Bayes-optimal value.
Within-domain tier conclusions hold with one exception: the ICU values
and measurement-count tiers swap order under the linear classifier.
The order of the mid-table domains also changes.

\section{The Temporal AUC Depends on the Time Offset}
\label{sec:offsets}

Offset curves in two domains measure the full tier at a
short, published, and long offset range. Drive temporal AUCs are
\BbOffShortAuc{}, \BbFullAuc{}, \BbOffLongAuc{} at 7--29, 30--90,
and 90--180 days; ICU temporal AUCs are \IcuOffShortAuc{}, \IcuFullAuc{},
\IcuOffLongAuc{} at 3--6, 6--24, and 24--48 hours. Both curves are
monotone and steep. Within a single domain the temporal AUC therefore
spans nearly as much as the whole cross-domain spread of
Table~\ref{tab:floors}, so a temporal AUC is a property of a dataset
\emph{together with a declared offset range}, and should be measured at offsets relevant to the intended use. The cross-unit placebo tracks both curves
in full (Appendix~\ref{sec:placebo}), showing that shared time structure is sufficient to reproduce
the increases.

Identity moves the other way and far less. ICU identity AUC at the
same three offset ranges is \RvIdOffShortIcuRawAuc{}, \RvIdRawAuc{}
and \RvIdOffLongIcuRawAuc{} on raw values with counts, and
\RvIdOffShortIcuLocfAuc{}, \RvIdLocfAuc{} and \RvIdOffLongIcuLocfAuc{}
under carry-forward. Drive identity AUC is \RvIdOffShortBbFullAuc{},
\IdBbFullAuc{} and \RvIdOffLongBbFullAuc{} at 7--29, 30--90 and
90--180 days, and \RvIdOffShortBbNoClocksAuc{}, \IdBbNoClocksAuc{}
and \RvIdOffLongBbNoClocksAuc{} without clock columns.

The second ICU dataset behaves the same way. On PhysioNet 2012 the
temporal AUC is \IcuBOffShortAuc{}, \IcuBFullAuc{} and
\IcuBOffLongAuc{} at 3--6, 6--24 and 24--40 hours (its records end at
48 hours, so the long range is shorter), while identity AUC
is \RvIdOffShortIcuBRawAuc{}, \RvIdIcuBRawAuc{} and
\RvIdOffLongIcuBRawAuc{} on raw values with counts and
\RvIdOffShortIcuBLocfAuc{}, \RvIdIcuBLocfAuc{} and
\RvIdOffLongIcuBLocfAuc{} under carry-forward.

\section{The Cross-Unit Placebo: Shared Structure versus
Within-Unit Signal}
\label{sec:placebo}

Under pooled scoring, label-1 rows are on average one offset later
than label-0 rows on the panel's clock, so a classifier can detect
structure shared across the population at matched clock positions
without learning anything about an individual unit.
The control is a cross-unit placebo. Draw the anchor
$(x_i, t_1)$ and the offset exactly as the real construction does,
but take the second window from a \emph{different} unit observed at
$t_2 = t_1{+}\delta$, label earlier/later, and score pooled as usual.
The pairing carries no within-unit information, so any signal it
detects is population structure. I compare its AUC with the within-unit
pairing AUC. Their difference is a diagnostic contrast, not an
additive statistical decomposition; AUC is nonlinear, and the
contrast may be negative.

\begin{table}[H]
\centering
\small
\begin{tabular}{llccc}
\toprule
Domain & Offsets & Temporal & Placebo & Difference \\
\midrule
Drives & 7--29 d & \PlcBbShortFloorAuc{} & \PlcBbShortAuc{} [\PlcBbShortCiLo, \PlcBbShortCiHi] & \PlcBbShortDiff{} \\
Drives & 30--90 d & \PlcBbMidFloorAuc{} & \PlcBbMidAuc{} [\PlcBbMidCiLo, \PlcBbMidCiHi] & \PlcBbMidDiff{} \\
Drives & 90--180 d & \PlcBbLongFloorAuc{} & \PlcBbLongAuc{} [\PlcBbLongCiLo, \PlcBbLongCiHi] & \PlcBbLongDiff{} \\
ICU & 3--6 h & \PlcIcuShortFloorAuc{} & \PlcIcuShortAuc{} [\PlcIcuShortCiLo, \PlcIcuShortCiHi] & \PlcIcuShortDiff{} \\
ICU & 6--24 h & \PlcIcuMidFloorAuc{} & \PlcIcuMidAuc{} [\PlcIcuMidCiLo, \PlcIcuMidCiHi] & \PlcIcuMidDiff{} \\
ICU & 24--48 h & \PlcIcuLongFloorAuc{} & \PlcIcuLongAuc{} [\PlcIcuLongCiLo, \PlcIcuLongCiHi] & \PlcIcuLongDiff{} \\
ICU (PhysioNet 2012) & 6--24 h & \IcuBFullAuc{} & \PlcIcuBMidAuc{} [\PlcIcuBMidCiLo, \PlcIcuBMidCiHi] & \PlcIcuBMidDiff{} \\
Mortgage (clocks) & 6--18 mo & \PlcFmFullFloorAuc{} & \PlcFmFullAuc{} [\PlcFmFullCiLo, \PlcFmFullCiHi] & \PlcFmFullDiff{} \\
Mortgage (no clocks) & 6--18 mo & \PlcFmNoClocksFloorAuc{} & \PlcFmNoClocksAuc{} [\PlcFmNoClocksCiLo, \PlcFmNoClocksCiHi] & \PlcFmNoClocksDiff{} \\
Mortgage (rate only) & 6--18 mo & \PlcFmRateFloorAuc{} & \PlcFmRateAuc{} [\PlcFmRateCiLo, \PlcFmRateCiHi] & \PlcFmRateDiff{} \\
Clickstreams & 14--60 d & \PlcOuladFloorAuc{} & \PlcOuladAuc{} [\PlcOuladCiLo, \PlcOuladCiHi] & \PlcOuladDiff{} \\
Wind & 14--60 d & \PlcWindFloorAuc{} & \PlcWindAuc{} [\PlcWindCiLo, \PlcWindCiHi] & \PlcWindDiff{} \\
Vessels & 14--60 d & \PlcMarFloorAuc{} & \PlcMarAuc{} [\PlcMarCiLo, \PlcMarCiHi] & \PlcMarDiff{} \\
\bottomrule
\end{tabular}
\caption{Cross-unit placebo beside the matched within-unit temporal
AUC (five-seed means, seed-0 unit-clustered bootstrap
intervals on the placebo; full tier except the labeled mortgage
variants). ``Difference'' is temporal AUC minus placebo
AUC, reported as a diagnostic contrast rather than an additive
within-unit component. The placebo reproduces most of the temporal AUC
everywhere except the mortgage clock tier, whose temporal structure is
unit-anchored; in the PhysioNet 2019 ICU 6--24 h, drive 90--180 d and mortgage
no-clock rows the temporal AUC sits above the placebo interval.}
\label{tab:placebo-iclr}
\end{table}

The placebo reproduces most of the temporal AUC in every measured cell
except mortgages with clock features (Table~\ref{tab:placebo-iclr}). In ICU records, matching stay
hours across patients retains the temporal discrimination; in drives,
matching calendar positions does the same. Shared time structure is
therefore sufficient to reproduce these scores. The mortgage clock
tier differs: cross-unit pairing lowers AUC by \PlcFmFullDiff{},
consistent with balance and age ordering records within a loan more
reliably than across loans. These comparisons identify shared
structure that can support discrimination; they do not isolate every
source of temporal information.

\section{Cohort Holdouts and Hospital-Matched Identity}
\label{app:extra-arms}

The controlled runs also measured a cohort arm not shown in the main
tables (Table~\ref{tab:cohort-arm}), and hospital-matched identity for
every ICU variant (Table~\ref{tab:site-identity}). In the \emph{cohort} arm, the test rows are half the units of
held-out cohorts (loan vintage for mortgages, drive model for drives,
hospital for ICU). Both training sets exclude the test units; overlap
training adds their sibling units from the same cohorts, and disjoint
training uses other cohorts only. Test rows and per-class training
counts are fixed as in Table~\ref{tab:bridge}. A positive $\Delta$
means that seeing other members of a test unit's cohort raises AUC.

\begin{table}[H]
\centering
\small
\setlength{\tabcolsep}{5pt}
\begin{tabular}{lllc}
\toprule
Experiment & Domain (cohort) & Tier & $\Delta$ cohort [range] \\
\midrule
Case-control & Mortgage (vintage) & full & \RvFmFullFamDelta{} [\RvFmFullFamLo, \RvFmFullFamHi] \\
Case-control & Mortgage (vintage) & no clocks & \RvFmNoClocksFamDelta{} [\RvFmNoClocksFamLo, \RvFmNoClocksFamHi] \\
Case-control & Drives (model) & full & \RvBbFullFamDelta{} [\RvBbFullFamLo, \RvBbFullFamHi] \\
Case-control & Drives (model) & no clocks & \RvBbNoClocksFamDelta{} [\RvBbNoClocksFamLo, \RvBbNoClocksFamHi] \\
Case-control & ICU (hospital) & carry-forward & \RvIcuLocfFamDelta{} [\RvIcuLocfFamLo, \RvIcuLocfFamHi] \\
Case-control & ICU (hospital) & carry-forward 6 h & \RvIcuLocfSixFamDelta{} [\RvIcuLocfSixFamLo, \RvIcuLocfSixFamHi] \\
Case-control & ICU (hospital) & model fill & \RvIcuModelTrainFamDelta{} [\RvIcuModelTrainFamLo, \RvIcuModelTrainFamHi] \\
Case-control & ICU (hospital) & raw & \RvIcuRawFamDelta{} [\RvIcuRawFamLo, \RvIcuRawFamHi] \\
Case-control & ICU (hospital) & mean fill & \RvIcuMeanTrainFamDelta{} [\RvIcuMeanTrainFamLo, \RvIcuMeanTrainFamHi] \\
Forecasting & ICU (hospital) & carry-forward & \RvFcLocfFamDelta{} [\RvFcLocfFamLo, \RvFcLocfFamHi] \\
Forecasting & ICU (hospital) & carry-forward 6 h & \RvFcLocfSixFamDelta{} [\RvFcLocfSixFamLo, \RvFcLocfSixFamHi] \\
Forecasting & ICU (hospital) & model fill & \RvFcModelTrainFamDelta{} [\RvFcModelTrainFamLo, \RvFcModelTrainFamHi] \\
Forecasting & ICU (hospital) & raw & \RvFcRawFamDelta{} [\RvFcRawFamLo, \RvFcRawFamHi] \\
Forecasting & ICU (hospital) & mean fill & \RvFcMeanTrainFamDelta{} [\RvFcMeanTrainFamLo, \RvFcMeanTrainFamHi] \\
\bottomrule
\end{tabular}
\caption{Cohort arm of the controlled experiments. Ranges are
empirical 2.5--97.5 percentile ranges over 200 replicates. ICU raw
includes the six-hour measurement counts, as in Table~\ref{tab:bridge}.}
\label{tab:cohort-arm}
\end{table}

Drives show a cohort effect that excludes zero in both tiers: training
on other drives of the same model raises AUC on held-out drives by
\RvBbFullFamDelta{}, and by \RvBbNoClocksFamDelta{} without clock
columns. Mortgage ranges include zero and are wide, because vintages
differ greatly in size and vintage is close to collinear with the
calendar. In the ICU, the hospital effect
is similar across imputations in both experiments: the forecasting
contrast between carry-forward and raw values is
\RvPairFcLocfRawFamDelta{} [\RvPairFcLocfRawFamLo,
\RvPairFcLocfRawFamHi], and the case-control contrast is \RvPairIcuLocfRawFamDelta{}
[\RvPairIcuLocfRawFamLo, \RvPairIcuLocfRawFamHi]. Both paired intervals include zero, so I
see no clear imputation effect on hospital composition, unlike the
clear effect on patient overlap.

\begin{table}[H]
\centering
\small
\begin{tabular}{lcc}
\toprule
ICU variant & All different-patient pairs & Same-hospital pairs \\
\midrule
carry-forward & \RvIdLocfAuc{} & \RvIdSiteLocfAuc{} [\RvIdSiteLocfLo, \RvIdSiteLocfHi] \\
carry-forward 6 h & \RvIdLocfSixAuc{} & \RvIdSiteLocfSixAuc{} [\RvIdSiteLocfSixLo, \RvIdSiteLocfSixHi] \\
model fill & \RvIdModelAuc{} & \RvIdSiteModelAuc{} [\RvIdSiteModelLo, \RvIdSiteModelHi] \\
raw & \RvIdRawAuc{} & \RvIdSiteRawAuc{} [\RvIdSiteRawLo, \RvIdSiteRawHi] \\
mean fill & \RvIdMeanAuc{} & \RvIdSiteMeanAuc{} [\RvIdSiteMeanLo, \RvIdSiteMeanHi] \\
counts only & \RvIdCountsAuc{} & \RvIdSiteCountsAuc{} [\RvIdSiteCountsLo, \RvIdSiteCountsHi] \\
\bottomrule
\end{tabular}
\caption{Identity AUC for every ICU variant, with different-patient
pairs drawn from any hospital or matched to the anchor's hospital.
Means over five seeds; brackets are intervals for the matched
column.}
\label{tab:site-identity}
\end{table}

Matching on hospital lowers identity AUC for every variant, most for
measurement counts alone and least under unlimited carry-forward, and
keeps the order of the variants: unlimited carry-forward, then the
model fill, then six-hour carry-forward, then raw values and mean fill,
then counts alone. The
ranking of preprocessing choices used in Section~\ref{sec:bridge}
therefore holds within a hospital.

\section{The Library on the Six Domains}
\label{app:library}

The \texttt{whowhen} 0.3.1 library implements corresponding identity
(re-identification) and timing (time-order) tests. Its classifier is
mojolearn 0.8.66 gradient boosting with symmetric trees
\citep{hendel2026mojolearn}, whereas the main-text harness uses LightGBM.
WhoWhen uses mojolearn's default \texttt{identical} mode for bitwise
reproducibility across supported Apple, NVIDIA and AMD GPUs and Apple
silicon (ARM64) and Linux x86-64 CPUs, given the same release, numerical
profile, inputs, hyperparameters and seed.
The library chooses offsets from the median gap between consecutive
observations up to half the median unit span. It uses three fixed
repeats, each holding out a quarter of the units; the harness uses five
random seeds with a fifth held out.

Library identity controls match on time, event status and automatically
detected unit-constant group columns. Here these include drive model and
capacity, mortgage vintage and ICU site. Identity sampling retains
event-bearing units; timing and placebo sampling use event-free units.
Pairs without eligible controls are dropped together. These changes in
sampling, offsets and classifier make the comparison descriptive, rather
than a controlled replication of the main-text estimates. The feature
sets are those of Table~\ref{tab:exposure}. The reported library results
cover the identity, timing, placebo and gap checks, with no forecast
horizon specified; they do not rerun the main-text forecasting experiments.

\begin{table}[H]
\centering
\small
\setlength{\tabcolsep}{4pt}
\begin{tabular}{llcccc}
\toprule
 & & \multicolumn{2}{c}{Identity} & \multicolumn{2}{c}{Timing} \\
\cmidrule(lr){3-4}\cmidrule(lr){5-6}
Domain (features) & Library offsets & Main text & Library & Library & Placebo \\
\midrule
Drives (all) & \WwBbGap & \IdBbFullAuc{} & \WwBbFullId{} & \WwBbFullTime{} & \WwBbFullPlac{} \\
Drives (no clocks) & \WwBbGap & \IdBbNoClocksAuc{} & \WwBbNoClocksId{} & \WwBbNoClocksTime{} & \WwBbNoClocksPlac{} \\
Mortgage (all) & \WwFmGap & \IdFmFullAuc{} & \WwFmFullId{} & \WwFmFullTime{} & \WwFmFullPlac{} \\
Mortgage (no clocks) & \WwFmGap & \IdFmNoClocksAuc{} & \WwFmNoClocksId{} & \WwFmNoClocksTime{} & \WwFmNoClocksPlac{} \\
ICU (carry-forward) & \WwIcuGap & \RvIdLocfAuc{} & \WwIcuLocfId{} & \WwIcuLocfTime{} & \WwIcuLocfPlac{} \\
ICU (no carry-forward) & \WwIcuGap & \RvIdRawAuc{} & \WwIcuRawId{} & \WwIcuRawTime{} & \WwIcuRawPlac{} \\
Vessels (all) & \WwMarGap & \IdMarFullAuc{} & \WwMarFullId{} & \WwMarFullTime{} & \WwMarFullPlac{} \\
Vessels (no location) & \WwMarGap & \IdMarNoLocAuc{} & \WwMarNoLocId{} & \WwMarNoLocTime{} & \WwMarNoLocPlac{} \\
Clickstreams (all) & \WwOuladGap & \IdOuladFullAuc{} & \WwOuladFullId{} & \WwOuladFullTime{} & \WwOuladFullPlac{} \\
Turbines (all) & \WwWindGap & --- & --- & \WwWindFullTime{} & \WwWindFullPlac{} \\
\bottomrule
\end{tabular}
\caption{WhoWhen 0.3.1 audit AUCs. \emph{Main text} repeats the
research-harness identity AUC of Table~\ref{tab:exposure}.
\emph{Placebo} uses a later row from a different unit observed at that
time. Means over three repeats. Turbine identity estimates are omitted
because twenty units are insufficient for reliable identity inference;
its timing estimates are exploratory.}
\label{tab:library}
\end{table}

The library shows high identity AUCs for drives, mortgages, vessels
and carried-forward ICU records, with lower values for raw ICU records
and clickstreams. These estimates use matched controls rather than the
nearest-neighbor construction of Table~\ref{tab:nearest}.
Timing AUCs are not comparable row for row with Table~\ref{tab:floors},
because they depend on the offset (Appendix~\ref{sec:offsets}) and the
sampling protocol. The placebo differs from timing AUC by at most
\WwMaxPlacDiff{} across the reported feature sets, and sometimes exceeds
it, consistent with substantial time structure shared across units.

Table~\ref{tab:library-gaps} splits each library offset range at its
midpoint. The largest short-to-long identity AUC drop is
\WwMaxIdGapDrop{}. Timing AUC is higher on the longer half in every
domain except vessels, where it is near chance on both.

\begin{table}[H]
\centering
\small
\begin{tabular}{lcccc}
\toprule
 & \multicolumn{2}{c}{Identity} & \multicolumn{2}{c}{Timing} \\
\cmidrule(lr){2-3}\cmidrule(lr){4-5}
Domain (all features) & Shorter half & Longer half & Shorter half & Longer half \\
\midrule
Drives & \WwBbFullIdShort{} & \WwBbFullIdLong{} & \WwBbFullTimeShort{} & \WwBbFullTimeLong{} \\
Mortgage & \WwFmFullIdShort{} & \WwFmFullIdLong{} & \WwFmFullTimeShort{} & \WwFmFullTimeLong{} \\
ICU (carry-forward) & \WwIcuLocfIdShort{} & \WwIcuLocfIdLong{} & \WwIcuLocfTimeShort{} & \WwIcuLocfTimeLong{} \\
ICU (no carry-forward) & \WwIcuRawIdShort{} & \WwIcuRawIdLong{} & \WwIcuRawTimeShort{} & \WwIcuRawTimeLong{} \\
Vessels & \WwMarFullIdShort{} & \WwMarFullIdLong{} & \WwMarFullTimeShort{} & \WwMarFullTimeLong{} \\
Clickstreams & \WwOuladFullIdShort{} & \WwOuladFullIdLong{} & \WwOuladFullTimeShort{} & \WwOuladFullTimeLong{} \\
Turbines & --- & --- & \WwWindFullTimeShort{} & \WwWindFullTimeLong{} \\
\bottomrule
\end{tabular}
\caption{Library AUCs on the shorter and the longer half of each
offset range of Table~\ref{tab:library}.}
\label{tab:library-gaps}
\end{table}

\section{Training-Size Control}
\label{app:bridge-submitted}

To measure sensitivity to training-set size, I keep the test rows
fixed and compare a fit on all remaining-unit training rows with a
fit on a random 80\% of those rows. The largest absolute mean AUC
difference across the controlled survey cells is \RvSizeMax{}.
The overlap comparisons in the main text match per-class training
counts between arms.

\section{Closest Methodological Comparisons}
\label{sec:relatedfull}

\textbf{Identity verification.} Biometric verification
\citep{jain2004introduction,odinaka2012ecg,monrose2000keystroke}
provides the same-unit versus different-unit classification task.
Patient re-identification from laboratory orders
\citep{johnson2019reidentification} is a close clinical precedent:
it ranks a day's orders against candidate patients to measure privacy
risk. Here, pairs are matched on the panel's time coordinate and
offset, and discrimination is compared with event-evaluation gaps.
Studies of identity confounding
\citep{tampu2022inflation,neto2019identity} measure the downstream
consequence; the audit measures recognizability before fitting an
event predictor.

\textbf{Temporal discrimination.} Classifier two-sample tests
\citep{lopezpaz2017revisiting,clemencon2009auc} provide the
methodological background.
Within-unit pairing also follows the self-matching principle of
case-crossover designs, whose time-trend corrections address changes
that survive matching \citep{maclure1991casecrossover,suissa1995casetimecontrol}.
The cross-unit placebo checks whether shared time structure is
sufficient for the observed discrimination. Unlike downstream
temporal-degradation studies \citep{vela2022temporal}, the audit
requires no fitted event model.

Temporal-order pretext tasks \citep{misra2016shuffle,wei2018arrow}
and instance discrimination \citep{wu2018instance} use related signals
for representation learning (Section~\ref{sec:related}). Their usefulness for training does not
make them evaluation errors: the deployment target and information
available at prediction determine whether using them is appropriate.

\end{document}

%% file: numbers_atlas.tex
\newcommand{\IcuFullAuc}{0.718}
\newcommand{\IcuFullCiLo}{0.706}
\newcommand{\IcuFullCiHi}{0.730}

\newcommand{\IcuValuesAuc}{0.677}

\newcommand{\IcuTextureAuc}{0.634}

\newcommand{\BbFullAuc}{0.598}

\newcommand{\BbNoClocksAuc}{0.527}

\newcommand{\BbTempAuc}{0.507}

\newcommand{\OuladFullAuc}{0.534}

\newcommand{\OuladTotalsAuc}{0.520}

\newcommand{\WindFullAuc}{0.539}

\newcommand{\WindNoClocksAuc}{0.540}

\newcommand{\WindAmbientAuc}{0.517}

\newcommand{\FmFullAuc}{0.834}

\newcommand{\FmNoClocksAuc}{0.549}

\newcommand{\FmRateAuc}{0.500}

\newcommand{\MarFullAuc}{0.527}
\newcommand{\MarFullCiLo}{0.496}
\newcommand{\MarFullCiHi}{0.558}

\newcommand{\MarNoLocAuc}{0.517}

\newcommand{\MarKinAuc}{0.505}

\newcommand{\LinIcuFullAuc}{0.636}

\newcommand{\LinBbFullAuc}{0.522}

\newcommand{\LinOuladFullAuc}{0.542}

\newcommand{\LinWindFullAuc}{0.536}

\newcommand{\LinFmFullAuc}{0.821}

\newcommand{\LinMarFullAuc}{0.506}

%% file: numbers_identity.tex
\newcommand{\IdFmFullAuc}{1.000}

\newcommand{\IdFmNoClocksAuc}{0.994}

\newcommand{\IdBbFullAuc}{1.000}

\newcommand{\IdBbNoClocksAuc}{0.998}

\newcommand{\IdIcuFullAuc}{0.993}

\newcommand{\IdIcuTextureAuc}{0.879}
\newcommand{\IdIcuTextureCiLo}{0.860}
\newcommand{\IdIcuTextureCiHi}{0.897}

\newcommand{\IdIcuRawFullAuc}{0.850}
\newcommand{\IdIcuRawFullCiLo}{0.828}
\newcommand{\IdIcuRawFullCiHi}{0.871}
\newcommand{\IdIcuRawFullPrimaryLo}{0.823}
\newcommand{\IdIcuRawFullPrimaryHi}{0.865}

\newcommand{\IdOuladFullAuc}{0.660}
\newcommand{\IdOuladFullCiLo}{0.624}
\newcommand{\IdOuladFullCiHi}{0.694}
\newcommand{\IdOuladFullPrimaryLo}{0.624}
\newcommand{\IdOuladFullPrimaryHi}{0.684}

\newcommand{\IdOuladTotalsAuc}{0.540}

\newcommand{\IdWindFullCiLo}{0.617}
\newcommand{\IdWindFullCiHi}{0.999}

\newcommand{\IdWindFullSeedLo}{0.747}
\newcommand{\IdWindFullSeedHi}{0.948}

\newcommand{\IdMarFullAuc}{0.955}
\newcommand{\IdMarFullCiLo}{0.932}
\newcommand{\IdMarFullCiHi}{0.973}

\newcommand{\IdMarNoLocAuc}{0.886}

\newcommand{\IdMarRepAuc}{0.732}

%% file: numbers_extras.tex
\newcommand{\BbOffShortAuc}{0.525}

\newcommand{\BbOffLongAuc}{0.734}

\newcommand{\IcuOffShortAuc}{0.592}

\newcommand{\IcuOffLongAuc}{0.881}

\newcommand{\IdBbCohortFullAuc}{0.996}
\newcommand{\IdBbCohortFullCiLo}{0.992}
\newcommand{\IdBbCohortFullCiHi}{0.998}

\newcommand{\IcuRetPanel}{40,336}

\newcommand{\BbRetPanel}{360,106}

\newcommand{\OuladRetPanel}{29,228}

\newcommand{\WindRetPanel}{20}

\newcommand{\FmRetPanel}{162,490}

\newcommand{\MarRetPanel}{6,248}
\newcommand{\MarRetOrdinary}{6,248}
\newcommand{\MarRetEligible}{4,628}

\newcommand{\RegenOuladCiLo}{0.639}
\newcommand{\RegenOuladCiHi}{0.674}

\newcommand{\RegenIcuRawCiLo}{0.836}
\newcommand{\RegenIcuRawCiHi}{0.859}

\newcommand{\IdWindPilotAuc}{0.917}

\newcommand{\RegenMeanGapMax}{0.004}
\newcommand{\PlcBbShortAuc}{0.544}
\newcommand{\PlcBbShortCiLo}{0.521}
\newcommand{\PlcBbShortCiHi}{0.565}

\newcommand{\PlcBbShortFloorAuc}{0.525}
\newcommand{\PlcBbShortDiff}{-0.020}
\newcommand{\PlcBbMidAuc}{0.592}
\newcommand{\PlcBbMidCiLo}{0.579}
\newcommand{\PlcBbMidCiHi}{0.605}

\newcommand{\PlcBbMidFloorAuc}{0.598}
\newcommand{\PlcBbMidDiff}{0.006}
\newcommand{\PlcBbLongAuc}{0.715}
\newcommand{\PlcBbLongCiLo}{0.702}
\newcommand{\PlcBbLongCiHi}{0.727}

\newcommand{\PlcBbLongFloorAuc}{0.734}
\newcommand{\PlcBbLongDiff}{0.019}
\newcommand{\PlcIcuShortAuc}{0.586}
\newcommand{\PlcIcuShortCiLo}{0.570}
\newcommand{\PlcIcuShortCiHi}{0.602}

\newcommand{\PlcIcuShortFloorAuc}{0.592}
\newcommand{\PlcIcuShortDiff}{0.005}
\newcommand{\PlcIcuMidAuc}{0.697}
\newcommand{\PlcIcuMidCiLo}{0.684}
\newcommand{\PlcIcuMidCiHi}{0.708}

\newcommand{\PlcIcuMidFloorAuc}{0.718}
\newcommand{\PlcIcuMidDiff}{0.021}
\newcommand{\PlcIcuLongAuc}{0.872}
\newcommand{\PlcIcuLongCiLo}{0.857}
\newcommand{\PlcIcuLongCiHi}{0.891}

\newcommand{\PlcIcuLongFloorAuc}{0.881}
\newcommand{\PlcIcuLongDiff}{0.009}
\newcommand{\PlcFmFullAuc}{0.736}
\newcommand{\PlcFmFullCiLo}{0.718}
\newcommand{\PlcFmFullCiHi}{0.753}

\newcommand{\PlcFmFullFloorAuc}{0.834}
\newcommand{\PlcFmFullDiff}{0.098}
\newcommand{\PlcFmNoClocksAuc}{0.529}
\newcommand{\PlcFmNoClocksCiLo}{0.516}
\newcommand{\PlcFmNoClocksCiHi}{0.542}

\newcommand{\PlcFmNoClocksFloorAuc}{0.549}
\newcommand{\PlcFmNoClocksDiff}{0.020}
\newcommand{\PlcFmRateAuc}{0.507}
\newcommand{\PlcFmRateCiLo}{0.490}
\newcommand{\PlcFmRateCiHi}{0.523}

\newcommand{\PlcFmRateFloorAuc}{0.500}
\newcommand{\PlcFmRateDiff}{-0.007}
\newcommand{\PlcOuladAuc}{0.555}
\newcommand{\PlcOuladCiLo}{0.542}
\newcommand{\PlcOuladCiHi}{0.567}

\newcommand{\PlcOuladFloorAuc}{0.534}
\newcommand{\PlcOuladDiff}{-0.021}
\newcommand{\PlcWindAuc}{0.563}
\newcommand{\PlcWindCiLo}{0.529}
\newcommand{\PlcWindCiHi}{0.599}

\newcommand{\PlcWindFloorAuc}{0.539}
\newcommand{\PlcWindDiff}{-0.024}
\newcommand{\PlcMarAuc}{0.546}
\newcommand{\PlcMarCiLo}{0.510}
\newcommand{\PlcMarCiHi}{0.580}

\newcommand{\PlcMarFloorAuc}{0.527}
\newcommand{\PlcMarDiff}{-0.019}

\newcommand{\BbStraddle}{95\%}
\newcommand{\FmStraddle}{75\%}
\newcommand{\OuladStraddle}{64\%}
\newcommand{\IcuStraddle}{52\%}

%% file: numbers_revision.tex
\newcommand{\RvIcuLocfFamDelta}{+0.085}
\newcommand{\RvIcuLocfFamLo}{+0.054}
\newcommand{\RvIcuLocfFamHi}{+0.115}

\newcommand{\RvIcuLocfGapDelta}{+0.060}

\newcommand{\RvIcuLocfTimeDelta}{+0.116}

\newcommand{\RvIcuLocfUnitDelta}{+0.120}
\newcommand{\RvIcuLocfUnitLo}{+0.108}
\newcommand{\RvIcuLocfUnitHi}{+0.131}
\newcommand{\RvIcuLocfUnitOv}{0.908}
\newcommand{\RvIcuLocfUnitDj}{0.788}
\newcommand{\RvIcuRawFamDelta}{+0.076}
\newcommand{\RvIcuRawFamLo}{+0.053}
\newcommand{\RvIcuRawFamHi}{+0.099}

\newcommand{\RvIcuRawGapDelta}{+0.007}

\newcommand{\RvIcuRawTimeDelta}{+0.013}

\newcommand{\RvIcuRawUnitDelta}{+0.017}
\newcommand{\RvIcuRawUnitLo}{+0.014}
\newcommand{\RvIcuRawUnitHi}{+0.020}
\newcommand{\RvIcuRawUnitOv}{0.769}
\newcommand{\RvIcuRawUnitDj}{0.752}

\newcommand{\RvIcuLocfSixFamDelta}{+0.070}
\newcommand{\RvIcuLocfSixFamLo}{+0.052}
\newcommand{\RvIcuLocfSixFamHi}{+0.087}

\newcommand{\RvIcuLocfSixGapDelta}{+0.011}

\newcommand{\RvIcuLocfSixTimeDelta}{+0.028}

\newcommand{\RvIcuLocfSixUnitDelta}{+0.057}
\newcommand{\RvIcuLocfSixUnitLo}{+0.051}
\newcommand{\RvIcuLocfSixUnitHi}{+0.063}
\newcommand{\RvIcuLocfSixUnitOv}{0.828}
\newcommand{\RvIcuLocfSixUnitDj}{0.770}
\newcommand{\RvIcuMeanTrainFamDelta}{+0.075}
\newcommand{\RvIcuMeanTrainFamLo}{+0.055}
\newcommand{\RvIcuMeanTrainFamHi}{+0.096}

\newcommand{\RvIcuMeanTrainGapDelta}{+0.007}

\newcommand{\RvIcuMeanTrainTimeDelta}{+0.013}

\newcommand{\RvIcuMeanTrainUnitDelta}{+0.018}
\newcommand{\RvIcuMeanTrainUnitLo}{+0.015}
\newcommand{\RvIcuMeanTrainUnitHi}{+0.021}
\newcommand{\RvIcuMeanTrainUnitOv}{0.769}
\newcommand{\RvIcuMeanTrainUnitDj}{0.751}
\newcommand{\RvIcuModelTrainFamDelta}{+0.096}
\newcommand{\RvIcuModelTrainFamLo}{+0.060}
\newcommand{\RvIcuModelTrainFamHi}{+0.159}

\newcommand{\RvIcuModelTrainGapDelta}{+0.024}

\newcommand{\RvIcuModelTrainTimeDelta}{+0.052}

\newcommand{\RvIcuModelTrainUnitDelta}{+0.053}
\newcommand{\RvIcuModelTrainUnitLo}{+0.047}
\newcommand{\RvIcuModelTrainUnitHi}{+0.060}
\newcommand{\RvIcuModelTrainUnitOv}{0.834}
\newcommand{\RvIcuModelTrainUnitDj}{0.781}
\newcommand{\RvPairIcuLocfRawFamDelta}{+0.009}
\newcommand{\RvPairIcuLocfRawFamLo}{-0.030}
\newcommand{\RvPairIcuLocfRawFamHi}{+0.048}

\newcommand{\RvPairIcuLocfRawUnitDelta}{+0.103}
\newcommand{\RvPairIcuLocfRawUnitLo}{+0.091}
\newcommand{\RvPairIcuLocfRawUnitHi}{+0.114}

\newcommand{\RvPairIcuModelTrainRawUnitDelta}{+0.036}

\newcommand{\RvPairIcuModelTrainLocfSixUnitDelta}{-0.004}
\newcommand{\RvPairIcuModelTrainLocfSixUnitLo}{-0.010}
\newcommand{\RvPairIcuModelTrainLocfSixUnitHi}{+0.002}

\newcommand{\RvFmFullFamDelta}{+0.070}
\newcommand{\RvFmFullFamLo}{-0.006}
\newcommand{\RvFmFullFamHi}{+0.216}

\newcommand{\RvFmFullGapDelta}{+0.032}
\newcommand{\RvFmFullGapLo}{+0.003}
\newcommand{\RvFmFullGapHi}{+0.065}

\newcommand{\RvFmFullTimeDelta}{+0.043}
\newcommand{\RvFmFullTimeLo}{+0.022}
\newcommand{\RvFmFullTimeHi}{+0.064}

\newcommand{\RvFmFullUnitDelta}{+0.061}
\newcommand{\RvFmFullUnitLo}{+0.046}
\newcommand{\RvFmFullUnitHi}{+0.080}
\newcommand{\RvFmFullUnitOv}{0.947}
\newcommand{\RvFmFullUnitDj}{0.886}
\newcommand{\RvFmNoClocksFamDelta}{+0.037}
\newcommand{\RvFmNoClocksFamLo}{-0.005}
\newcommand{\RvFmNoClocksFamHi}{+0.111}

\newcommand{\RvFmNoClocksGapDelta}{+0.027}

\newcommand{\RvFmNoClocksTimeDelta}{+0.064}
\newcommand{\RvFmNoClocksTimeLo}{+0.049}
\newcommand{\RvFmNoClocksTimeHi}{+0.081}

\newcommand{\RvFmNoClocksUnitDelta}{+0.046}
\newcommand{\RvFmNoClocksUnitLo}{+0.037}
\newcommand{\RvFmNoClocksUnitHi}{+0.059}
\newcommand{\RvFmNoClocksUnitOv}{0.812}
\newcommand{\RvFmNoClocksUnitDj}{0.766}

\newcommand{\RvOuladFullGapDelta}{+0.002}

\newcommand{\RvOuladFullTimeDelta}{+0.003}

\newcommand{\RvOuladFullUnitDelta}{+0.002}
\newcommand{\RvOuladFullUnitLo}{-0.003}
\newcommand{\RvOuladFullUnitHi}{+0.006}
\newcommand{\RvOuladFullUnitOv}{0.588}
\newcommand{\RvOuladFullUnitDj}{0.586}
\newcommand{\RvOuladTotalsGapDelta}{+0.000}

\newcommand{\RvOuladTotalsTimeDelta}{+0.000}

\newcommand{\RvOuladTotalsUnitDelta}{+0.000}
\newcommand{\RvOuladTotalsUnitLo}{-0.004}
\newcommand{\RvOuladTotalsUnitHi}{+0.003}
\newcommand{\RvOuladTotalsUnitOv}{0.540}
\newcommand{\RvOuladTotalsUnitDj}{0.540}

\newcommand{\RvBbFullFamDelta}{+0.105}
\newcommand{\RvBbFullFamLo}{+0.037}
\newcommand{\RvBbFullFamHi}{+0.219}

\newcommand{\RvBbFullGapDelta}{+0.019}
\newcommand{\RvBbFullGapLo}{-0.006}
\newcommand{\RvBbFullGapHi}{+0.044}

\newcommand{\RvBbFullTimeDelta}{+0.041}
\newcommand{\RvBbFullTimeLo}{+0.022}
\newcommand{\RvBbFullTimeHi}{+0.062}

\newcommand{\RvBbFullUnitDelta}{+0.045}
\newcommand{\RvBbFullUnitLo}{+0.038}
\newcommand{\RvBbFullUnitHi}{+0.053}
\newcommand{\RvBbFullUnitOv}{0.913}
\newcommand{\RvBbFullUnitDj}{0.867}
\newcommand{\RvBbNoClocksFamDelta}{+0.048}
\newcommand{\RvBbNoClocksFamLo}{+0.008}
\newcommand{\RvBbNoClocksFamHi}{+0.100}

\newcommand{\RvBbNoClocksGapDelta}{+0.021}

\newcommand{\RvBbNoClocksTimeDelta}{+0.038}
\newcommand{\RvBbNoClocksTimeLo}{+0.024}
\newcommand{\RvBbNoClocksTimeHi}{+0.055}

\newcommand{\RvBbNoClocksUnitDelta}{+0.045}
\newcommand{\RvBbNoClocksUnitLo}{+0.037}
\newcommand{\RvBbNoClocksUnitHi}{+0.053}
\newcommand{\RvBbNoClocksUnitOv}{0.874}
\newcommand{\RvBbNoClocksUnitDj}{0.829}
\newcommand{\RvSizeMax}{0.0022}

\newcommand{\RvMarFullUnitDelta}{+0.019}
\newcommand{\RvMarFullUnitLo}{+0.006}
\newcommand{\RvMarFullUnitHi}{+0.030}
\newcommand{\RvMarFullUnitOv}{0.884}
\newcommand{\RvMarFullUnitDj}{0.865}
\newcommand{\RvMarFullTimeDelta}{+0.014}
\newcommand{\RvMarFullTimeLo}{-0.010}
\newcommand{\RvMarFullTimeHi}{+0.038}

\newcommand{\RvMarFullGapDelta}{+0.013}
\newcommand{\RvMarFullGapLo}{-0.016}
\newcommand{\RvMarFullGapHi}{+0.047}

\newcommand{\RvMarNoLocUnitDelta}{+0.012}
\newcommand{\RvMarNoLocUnitLo}{-0.001}
\newcommand{\RvMarNoLocUnitHi}{+0.025}
\newcommand{\RvMarNoLocUnitOv}{0.836}
\newcommand{\RvMarNoLocUnitDj}{0.824}
\newcommand{\RvMarNoLocTimeDelta}{+0.009}

\newcommand{\RvMarNoLocGapDelta}{+0.007}

\newcommand{\RvIcuMeanFillMaxDiff}{0.0002}

\newcommand{\RvFcPosUnits}{2{,}506}
\newcommand{\RvFcHorizon}{12}

\newcommand{\RvFcCensored}{426}
\newcommand{\RvFcLocfUnitDelta}{+0.147}
\newcommand{\RvFcLocfUnitLo}{+0.133}
\newcommand{\RvFcLocfUnitHi}{+0.162}
\newcommand{\RvFcLocfUnitOv}{0.887}
\newcommand{\RvFcLocfUnitDj}{0.740}
\newcommand{\RvFcLocfFamDelta}{+0.054}
\newcommand{\RvFcLocfFamLo}{+0.023}
\newcommand{\RvFcLocfFamHi}{+0.097}

\newcommand{\RvFcRawUnitDelta}{+0.031}
\newcommand{\RvFcRawUnitLo}{+0.025}
\newcommand{\RvFcRawUnitHi}{+0.038}
\newcommand{\RvFcRawUnitOv}{0.724}
\newcommand{\RvFcRawUnitDj}{0.694}
\newcommand{\RvFcRawFamDelta}{+0.074}
\newcommand{\RvFcRawFamLo}{+0.053}
\newcommand{\RvFcRawFamHi}{+0.095}

\newcommand{\RvFcLocfSixUnitDelta}{+0.110}
\newcommand{\RvFcLocfSixUnitLo}{+0.097}
\newcommand{\RvFcLocfSixUnitHi}{+0.121}
\newcommand{\RvFcLocfSixUnitOv}{0.816}
\newcommand{\RvFcLocfSixUnitDj}{0.706}
\newcommand{\RvFcLocfSixFamDelta}{+0.059}
\newcommand{\RvFcLocfSixFamLo}{+0.031}
\newcommand{\RvFcLocfSixFamHi}{+0.087}

\newcommand{\RvFcMeanTrainUnitDelta}{+0.032}
\newcommand{\RvFcMeanTrainUnitLo}{+0.026}
\newcommand{\RvFcMeanTrainUnitHi}{+0.039}
\newcommand{\RvFcMeanTrainUnitOv}{0.725}
\newcommand{\RvFcMeanTrainUnitDj}{0.693}
\newcommand{\RvFcMeanTrainFamDelta}{+0.072}
\newcommand{\RvFcMeanTrainFamLo}{+0.052}
\newcommand{\RvFcMeanTrainFamHi}{+0.091}

\newcommand{\RvFcModelTrainUnitDelta}{+0.074}
\newcommand{\RvFcModelTrainUnitLo}{+0.065}
\newcommand{\RvFcModelTrainUnitHi}{+0.083}
\newcommand{\RvFcModelTrainUnitOv}{0.808}
\newcommand{\RvFcModelTrainUnitDj}{0.734}
\newcommand{\RvFcModelTrainFamDelta}{+0.080}
\newcommand{\RvFcModelTrainFamLo}{+0.056}
\newcommand{\RvFcModelTrainFamHi}{+0.116}

\newcommand{\RvFcLocfRawOvLead}{0.163}
\newcommand{\RvFcLocfRawDjLead}{0.046}
\newcommand{\RvFcLocfModelOvLead}{0.079}
\newcommand{\RvFcLocfModelDjLead}{0.006}
\newcommand{\RvFcModelLocfSixOvLead}{-0.008}
\newcommand{\RvFcModelLocfSixDjLead}{+0.028}
\newcommand{\RvPairFcLocfRawUnitDelta}{+0.117}
\newcommand{\RvPairFcLocfRawUnitLo}{+0.102}
\newcommand{\RvPairFcLocfRawUnitHi}{+0.131}
\newcommand{\RvPairFcLocfRawFamDelta}{-0.019}
\newcommand{\RvPairFcLocfRawFamLo}{-0.045}
\newcommand{\RvPairFcLocfRawFamHi}{+0.015}

\newcommand{\RvPairFcModelTrainLocfSixUnitDelta}{-0.035}
\newcommand{\RvPairFcModelTrainLocfSixUnitLo}{-0.048}
\newcommand{\RvPairFcModelTrainLocfSixUnitHi}{-0.022}

\newcommand{\RvIdLocfAuc}{0.993}

\newcommand{\RvIdRawAuc}{0.916}

\newcommand{\RvIdMeanAuc}{0.914}

\newcommand{\RvIdLocfSixAuc}{0.946}

\newcommand{\RvIdCountsAuc}{0.879}

\newcommand{\RvIdSiteLocfAuc}{0.989}
\newcommand{\RvIdSiteLocfLo}{0.984}
\newcommand{\RvIdSiteLocfHi}{0.992}
\newcommand{\RvIdSiteRawAuc}{0.879}
\newcommand{\RvIdSiteRawLo}{0.859}
\newcommand{\RvIdSiteRawHi}{0.899}
\newcommand{\RvIdSiteMeanAuc}{0.876}
\newcommand{\RvIdSiteMeanLo}{0.852}
\newcommand{\RvIdSiteMeanHi}{0.896}
\newcommand{\RvIdSiteLocfSixAuc}{0.920}
\newcommand{\RvIdSiteLocfSixLo}{0.905}
\newcommand{\RvIdSiteLocfSixHi}{0.933}
\newcommand{\RvIdSiteCountsAuc}{0.820}
\newcommand{\RvIdSiteCountsLo}{0.791}
\newcommand{\RvIdSiteCountsHi}{0.847}
\newcommand{\RvIdModelAuc}{0.978}

\newcommand{\RvIdSiteModelAuc}{0.971}
\newcommand{\RvIdSiteModelLo}{0.963}
\newcommand{\RvIdSiteModelHi}{0.978}
\newcommand{\RvIdOffShortBbFullAuc}{1.000}

\newcommand{\RvIdOffShortBbNoClocksAuc}{0.999}

\newcommand{\RvIdOffLongBbFullAuc}{1.000}

\newcommand{\RvIdOffLongBbNoClocksAuc}{0.998}

\newcommand{\RvIdOffShortIcuLocfAuc}{0.997}

\newcommand{\RvIdOffShortIcuRawAuc}{0.963}

\newcommand{\RvIdOffLongIcuLocfAuc}{0.979}

\newcommand{\RvIdOffLongIcuRawAuc}{0.860}

\newcommand{\RvSepCells}{13}
\newcommand{\RvSepHighN}{10}

\newcommand{\RvSepLowN}{3}

\newcommand{\RvChRawUnitDelta}{+0.034}
\newcommand{\RvChRawUnitLo}{+0.028}
\newcommand{\RvChRawUnitHi}{+0.040}

\newcommand{\RvChRawUnitDj}{0.681}
\newcommand{\RvChRawChronoDelta}{-0.004}
\newcommand{\RvChRawChronoLo}{-0.008}
\newcommand{\RvChRawChronoHi}{-0.000}

\newcommand{\RvChLocfSixUnitDelta}{+0.120}
\newcommand{\RvChLocfSixUnitLo}{+0.108}
\newcommand{\RvChLocfSixUnitHi}{+0.132}

\newcommand{\RvChLocfSixChronoDelta}{-0.006}
\newcommand{\RvChLocfSixChronoLo}{-0.012}
\newcommand{\RvChLocfSixChronoHi}{-0.000}

\newcommand{\RvChLocfUnitDelta}{+0.151}
\newcommand{\RvChLocfUnitLo}{+0.138}
\newcommand{\RvChLocfUnitHi}{+0.164}

\newcommand{\RvChLocfUnitDj}{0.732}
\newcommand{\RvChLocfChronoDelta}{-0.007}
\newcommand{\RvChLocfChronoLo}{-0.014}
\newcommand{\RvChLocfChronoHi}{+0.001}

\newcommand{\RvChMeanTrainUnitDelta}{+0.035}
\newcommand{\RvChMeanTrainUnitLo}{+0.029}
\newcommand{\RvChMeanTrainUnitHi}{+0.040}

\newcommand{\RvChMeanTrainChronoDelta}{-0.004}
\newcommand{\RvChMeanTrainChronoLo}{-0.007}
\newcommand{\RvChMeanTrainChronoHi}{+0.000}

\newcommand{\RvChTimeOnlyUnitDj}{0.728}

\newcommand{\RvChCountsTimeUnitDj}{0.761}

\newcommand{\RvChRawTimeUnitDj}{0.779}

\newcommand{\RvPairChLocfRawUnitDelta}{+0.117}
\newcommand{\RvPairChLocfRawUnitLo}{+0.104}
\newcommand{\RvPairChLocfRawUnitHi}{+0.130}

\newcommand{\RvChLeadLocfRawOv}{+0.167}

\newcommand{\RvChLeadLocfRawDj}{+0.050}

\newcommand{\RvChLeadTimeOnlyRawDj}{+0.046}
\newcommand{\RvChLeadTimeOnlyRawDjLo}{+0.023}
\newcommand{\RvChLeadTimeOnlyRawDjHi}{+0.070}

\newcommand{\RvChLeadCountsTimeLocfDj}{+0.029}
\newcommand{\RvChLeadCountsTimeLocfDjLo}{+0.013}
\newcommand{\RvChLeadCountsTimeLocfDjHi}{+0.047}

\newcommand{\RvChLeadRawTimeCountsTimeDj}{+0.018}
\newcommand{\RvChLeadRawTimeCountsTimeDjLo}{+0.008}
\newcommand{\RvChLeadRawTimeCountsTimeDjHi}{+0.026}

\newcommand{\RvChTestPatients}{1,305}
\newcommand{\RvChContribPatients}{805}
\newcommand{\RvIcuBRawUnitDelta}{+0.049}
\newcommand{\RvIcuBRawUnitLo}{+0.042}
\newcommand{\RvIcuBRawUnitHi}{+0.057}
\newcommand{\RvIcuBRawUnitOv}{0.796}
\newcommand{\RvIcuBRawUnitDj}{0.746}
\newcommand{\RvIcuBMeanTrainUnitDelta}{+0.050}
\newcommand{\RvIcuBMeanTrainUnitLo}{+0.042}
\newcommand{\RvIcuBMeanTrainUnitHi}{+0.059}
\newcommand{\RvIcuBMeanTrainUnitOv}{0.795}
\newcommand{\RvIcuBMeanTrainUnitDj}{0.745}
\newcommand{\RvIcuBLocfSixUnitDelta}{+0.071}
\newcommand{\RvIcuBLocfSixUnitLo}{+0.063}
\newcommand{\RvIcuBLocfSixUnitHi}{+0.080}
\newcommand{\RvIcuBLocfSixUnitOv}{0.869}
\newcommand{\RvIcuBLocfSixUnitDj}{0.797}
\newcommand{\RvIcuBLocfUnitDelta}{+0.132}
\newcommand{\RvIcuBLocfUnitLo}{+0.117}
\newcommand{\RvIcuBLocfUnitHi}{+0.146}
\newcommand{\RvIcuBLocfUnitOv}{0.946}
\newcommand{\RvIcuBLocfUnitDj}{0.815}
\newcommand{\RvPairIcuBLocfRawUnitDelta}{+0.083}
\newcommand{\RvPairIcuBLocfRawUnitLo}{+0.068}
\newcommand{\RvPairIcuBLocfRawUnitHi}{+0.099}
\newcommand{\RvPairIcuBLocfSixRawUnitDelta}{+0.022}
\newcommand{\RvPairIcuBLocfSixRawUnitLo}{+0.014}
\newcommand{\RvPairIcuBLocfSixRawUnitHi}{+0.031}

\newcommand{\RvPairIcuBMeanTrainRawUnitDelta}{+0.001}
\newcommand{\RvPairIcuBMeanTrainRawUnitLo}{-0.005}
\newcommand{\RvPairIcuBMeanTrainRawUnitHi}{+0.006}
\newcommand{\RvIcuBLocfRawOvLead}{0.150}
\newcommand{\RvIcuBLocfRawDjLead}{0.068}
\newcommand{\RvIdIcuBLocfAuc}{0.998}
\newcommand{\RvIdIcuBRawAuc}{0.974}
\newcommand{\RvIdIcuBMeanAuc}{0.972}
\newcommand{\RvIdIcuBLocfSixAuc}{0.985}
\newcommand{\RvIdIcuBCountsAuc}{0.939}
\newcommand{\IcuBFullAuc}{0.662}

\newcommand{\IcuBValuesAuc}{0.638}

\newcommand{\IcuBTextureAuc}{0.595}
\newcommand{\IcuBTextureCiLo}{0.572}
\newcommand{\IcuBTextureCiHi}{0.619}
\newcommand{\PlcIcuBMidAuc}{0.659}
\newcommand{\PlcIcuBMidCiLo}{0.641}
\newcommand{\PlcIcuBMidCiHi}{0.679}
\newcommand{\PlcIcuBMidDiff}{0.003}
\newcommand{\IcuBOffShortAuc}{0.556}
\newcommand{\RvIdOffShortIcuBLocfAuc}{1.000}
\newcommand{\RvIdOffShortIcuBRawAuc}{0.993}
\newcommand{\IcuBOffLongAuc}{0.883}
\newcommand{\RvIdOffLongIcuBLocfAuc}{0.993}
\newcommand{\RvIdOffLongIcuBRawAuc}{0.939}
\newcommand{\RvStARawOv}{0.742}
\newcommand{\RvStARawDj}{0.697}
\newcommand{\RvStARawDelta}{+0.045}
\newcommand{\RvStARawLo}{+0.036}
\newcommand{\RvStARawHi}{+0.055}
\newcommand{\RvStARawChDelta}{-0.003}
\newcommand{\RvStARawChLo}{-0.008}
\newcommand{\RvStARawChHi}{-0.000}
\newcommand{\RvStALocfSixOv}{0.849}
\newcommand{\RvStALocfSixDj}{0.697}
\newcommand{\RvStALocfSixDelta}{+0.151}
\newcommand{\RvStALocfSixLo}{+0.135}
\newcommand{\RvStALocfSixHi}{+0.169}
\newcommand{\RvStALocfSixChDelta}{-0.005}
\newcommand{\RvStALocfSixChLo}{-0.014}
\newcommand{\RvStALocfSixChHi}{+0.001}
\newcommand{\RvStALocfOv}{0.923}
\newcommand{\RvStALocfDj}{0.741}
\newcommand{\RvStALocfDelta}{+0.182}
\newcommand{\RvStALocfLo}{+0.162}
\newcommand{\RvStALocfHi}{+0.200}
\newcommand{\RvStALocfChDelta}{-0.007}
\newcommand{\RvStALocfChLo}{-0.016}
\newcommand{\RvStALocfChHi}{+0.002}
\newcommand{\RvStAPairLocfRawDelta}{+0.138}
\newcommand{\RvStAPairLocfRawLo}{+0.117}
\newcommand{\RvStAPairLocfRawHi}{+0.158}
\newcommand{\RvStATimeOnlyDj}{0.730}
\newcommand{\RvStALeadTimeOnlyRaw}{+0.033}
\newcommand{\RvStALeadTimeOnlyRawLo}{+0.006}
\newcommand{\RvStALeadTimeOnlyRawHi}{+0.062}
\newcommand{\RvStBRawOv}{0.745}
\newcommand{\RvStBRawDj}{0.698}
\newcommand{\RvStBRawDelta}{+0.047}
\newcommand{\RvStBRawLo}{+0.036}
\newcommand{\RvStBRawHi}{+0.060}
\newcommand{\RvStBRawChDelta}{-0.004}
\newcommand{\RvStBRawChLo}{-0.009}
\newcommand{\RvStBRawChHi}{+0.002}
\newcommand{\RvStBLocfSixOv}{0.864}
\newcommand{\RvStBLocfSixDj}{0.711}
\newcommand{\RvStBLocfSixDelta}{+0.153}
\newcommand{\RvStBLocfSixLo}{+0.130}
\newcommand{\RvStBLocfSixHi}{+0.177}
\newcommand{\RvStBLocfSixChDelta}{-0.006}
\newcommand{\RvStBLocfSixChLo}{-0.016}
\newcommand{\RvStBLocfSixChHi}{+0.002}
\newcommand{\RvStBLocfOv}{0.942}
\newcommand{\RvStBLocfDj}{0.756}
\newcommand{\RvStBLocfDelta}{+0.187}
\newcommand{\RvStBLocfLo}{+0.165}
\newcommand{\RvStBLocfHi}{+0.206}
\newcommand{\RvStBLocfChDelta}{-0.013}
\newcommand{\RvStBLocfChLo}{-0.025}
\newcommand{\RvStBLocfChHi}{-0.000}
\newcommand{\RvStBPairLocfRawDelta}{+0.140}
\newcommand{\RvStBPairLocfRawLo}{+0.118}
\newcommand{\RvStBPairLocfRawHi}{+0.163}
\newcommand{\RvStBTimeOnlyDj}{0.771}
\newcommand{\RvStBLeadTimeOnlyRaw}{+0.073}
\newcommand{\RvStBLeadTimeOnlyRawLo}{+0.034}
\newcommand{\RvStBLeadTimeOnlyRawHi}{+0.114}

%% file: numbers_library.tex
\newcommand{\NrBbFullAuc}{0.993}
\newcommand{\NrBbFullCiLo}{0.987}
\newcommand{\NrBbFullCiHi}{0.997}

\newcommand{\NrFmFullAuc}{0.998}
\newcommand{\NrFmFullCiLo}{0.995}
\newcommand{\NrFmFullCiHi}{0.999}

\newcommand{\NrIcuFullAuc}{0.977}
\newcommand{\NrIcuFullCiLo}{0.969}
\newcommand{\NrIcuFullCiHi}{0.983}
\newcommand{\NrIcuTextureAuc}{0.685}

\newcommand{\NrIcuVarRawAuc}{0.790}
\newcommand{\NrIcuVarRawCiLo}{0.766}
\newcommand{\NrIcuVarRawCiHi}{0.814}

\newcommand{\NrMarFullAuc}{0.923}
\newcommand{\NrMarFullCiLo}{0.894}
\newcommand{\NrMarFullCiHi}{0.947}

\newcommand{\NrOuladFullAuc}{0.542}
\newcommand{\NrOuladFullCiLo}{0.504}
\newcommand{\NrOuladFullCiHi}{0.577}

\newcommand{\WwIcuGap}{1--18.5 h}
\newcommand{\WwIcuRawId}{0.921}

\newcommand{\WwIcuRawTime}{0.602}
\newcommand{\WwIcuRawPlac}{0.590}
\newcommand{\WwIcuRawIdShort}{0.943}
\newcommand{\WwIcuRawIdLong}{0.870}
\newcommand{\WwIcuRawTimeShort}{0.578}
\newcommand{\WwIcuRawTimeLong}{0.680}
\newcommand{\WwIcuLocfId}{0.992}

\newcommand{\WwIcuLocfTime}{0.637}
\newcommand{\WwIcuLocfPlac}{0.621}
\newcommand{\WwIcuLocfIdShort}{0.994}
\newcommand{\WwIcuLocfIdLong}{0.985}
\newcommand{\WwIcuLocfTimeShort}{0.595}
\newcommand{\WwIcuLocfTimeLong}{0.754}
\newcommand{\WwBbGap}{1--227 d}
\newcommand{\WwBbFullId}{0.991}

\newcommand{\WwBbFullTime}{0.631}
\newcommand{\WwBbFullPlac}{0.638}
\newcommand{\WwBbFullIdShort}{0.994}
\newcommand{\WwBbFullIdLong}{0.992}
\newcommand{\WwBbFullTimeShort}{0.579}
\newcommand{\WwBbFullTimeLong}{0.781}
\newcommand{\WwBbNoClocksId}{0.980}

\newcommand{\WwBbNoClocksTime}{0.542}
\newcommand{\WwBbNoClocksPlac}{0.552}

\newcommand{\WwFmGap}{31--242.5 d}
\newcommand{\WwFmFullId}{1.000}

\newcommand{\WwFmFullTime}{0.673}
\newcommand{\WwFmFullPlac}{0.664}
\newcommand{\WwFmFullIdShort}{1.000}
\newcommand{\WwFmFullIdLong}{1.000}
\newcommand{\WwFmFullTimeShort}{0.629}
\newcommand{\WwFmFullTimeLong}{0.756}
\newcommand{\WwFmNoClocksId}{0.996}

\newcommand{\WwFmNoClocksTime}{0.525}
\newcommand{\WwFmNoClocksPlac}{0.534}

\newcommand{\WwMarGap}{11--132.5 d}
\newcommand{\WwMarFullId}{0.956}

\newcommand{\WwMarFullTime}{0.518}
\newcommand{\WwMarFullPlac}{0.574}
\newcommand{\WwMarFullIdShort}{0.960}
\newcommand{\WwMarFullIdLong}{0.949}
\newcommand{\WwMarFullTimeShort}{0.527}
\newcommand{\WwMarFullTimeLong}{0.522}
\newcommand{\WwMarNoLocId}{0.891}

\newcommand{\WwMarNoLocTime}{0.512}
\newcommand{\WwMarNoLocPlac}{0.571}

\newcommand{\WwOuladGap}{1--117 d}
\newcommand{\WwOuladFullId}{0.676}

\newcommand{\WwOuladFullTime}{0.546}
\newcommand{\WwOuladFullPlac}{0.568}
\newcommand{\WwOuladFullIdShort}{0.679}
\newcommand{\WwOuladFullIdLong}{0.663}
\newcommand{\WwOuladFullTimeShort}{0.535}
\newcommand{\WwOuladFullTimeLong}{0.591}
\newcommand{\WwWindGap}{1--182 d}

\newcommand{\WwWindFullTime}{0.566}
\newcommand{\WwWindFullPlac}{0.556}

\newcommand{\WwWindFullTimeShort}{0.552}
\newcommand{\WwWindFullTimeLong}{0.732}
\newcommand{\WwMaxIdGapDrop}{0.073}

\newcommand{\WwMaxPlacDiff}{0.060}

%% file: identity_balance.tex
\begin{tabular}{lrrrr}
\toprule
Panel & Train same & Test same & Test different & Test unmatched \\
\midrule
Drives & 8,000 & 2,000 & 2,000 & 0 \\
Drives (cohort matched) & 8,000--8,002 & 2,001 & 1,999--2,000 & 1--2 \\
Mortgage & 8,000 & 2,000 & 2,000 & 0 \\
Vessels & 3,200 & 800 & 800 & 0 \\
Clickstreams & 8,000 & 2,000 & 2,000 & 0 \\
ICU (carry-forward) & 8,000 & 2,000 & 2,000 & 0 \\
ICU (raw) & 8,000 & 2,000 & 2,000 & 0 \\
Wind & 251--335 & 58--137 & 58--122 & 0--15 \\
\bottomrule
\end{tabular}